\documentclass[letterpaper, final, 10 pt, conference]{ieeeconf}  % Comment this line out if you need a4paper

\usepackage{hyperref}
\usepackage{graphicx} % Allows including images
\usepackage{subfiles}
\usepackage{multirow}
\usepackage[ruled,vlined]{algorithm2e}
\usepackage{here}
\usepackage{amsfonts}
\usepackage{amsmath}
\usepackage{setspace}
\usepackage{multirow}
\usepackage{booktabs, tabularx} % Allows the use of \toprule, \midrule and \bottomrule in tables
\usepackage[font=small]{caption}
\usepackage{subcaption}
\usepackage{algpseudocode}% http://ctan.org/pkg/algorithmicx
\usepackage{color}
\usepackage[
    style=ieee,
    natbib=true,
    citestyle=numeric-comp,
    doi=false,
    isbn=false,
    url=false]{biblatex} 
\AtEveryBibitem{%
  \clearfield{note}%
}

\IEEEoverridecommandlockouts                              % This command is only needed if 
\title{\LARGE \bf
OPIRL: Sample Efficient Off-Policy Inverse Reinforcement Learning via Distribution Matching
}

\author{
    Hana Hoshino$^{1}$, Kei Ota$^{1, 2}$, Asako Kanezaki$^{1}$ and Rio Yokota$^{3}$%
\thanks{$^{1}$ School of Computing, Department of Computer Science, Tokyo Institute of Technology, Japan.}%
\thanks{$^{2}$ Information Technology R\&D Center, Mitsubishi Electric Corporation, Japan.}
\thanks{$^{3}$ Global Scientific Information and Computing Center, Tokyo Institute of Technology, Japan.}
\thanks{This work is supported by JST CREST Grant Number JPMJCR19F5.}}

\newcommand{\action}{\mathcal{A}}
\newcommand{\state}{\mathcal{S}}
\newcommand{\distagent}{\ensuremath{\rho^{\pi_\theta}}}
\newcommand{\distrb}{\ensuremath{\rho^{R}}}
\newcommand{\distexp}{\ensuremath{\rho^{\mathrm{exp}}}}
\newcommand{\fdivergence}{{\it f}-divergence }

\DeclareMathOperator*{\argmax}{arg\,max}

\newcommand{\fix}[1]{{\color{black} #1}}

\begin{document}

\maketitle
\thispagestyle{empty}
\pagestyle{empty}

%%%%%%%%%%%%%%%%%%%%%%%%%%%%%%%%%%%%%%%%%%%%%%%%%%%%%%%%%%%%%%%%%%%%%%%%%%%%%%%%
\begin{abstract}

% significant
Inverse Reinforcement Learning (IRL) is attractive in scenarios where reward engineering can be tedious.
However, prior IRL algorithms use on-policy transitions, which require intensive sampling from the current policy for stable and optimal performance.
This limits IRL applications in the real world, where environment interactions can become highly expensive.
To tackle this problem, we present Off-Policy Inverse Reinforcement Learning (OPIRL), which (1) adopts off-policy data distribution instead of on-policy and enables significant reduction of the number of interactions with the environment, (2) learns a reward function that is transferable with high generalization capabilities on changing dynamics, and (3) leverages mode-covering behavior for faster convergence.
We demonstrate that our method is considerably more sample efficient and generalizes to novel environments through the experiments.
Our method achieves better or comparable results on policy performance baselines with significantly fewer interactions.
Furthermore, we empirically show that the recovered reward function generalizes to different tasks where prior arts are prone to fail.
\end{abstract}

\begin{keywords}
Imitation Learning, Transfer Learning, Learning from Demonstration, Inverse Reinforcement Learning.
\end{keywords}

% Introduction %%%%%%%%%%%%%%%%%%%%%%%%%%%%%%%%%%%%%%%%%%%%%%%%%%%%%%%%%%%%%%%%%%%%%%%%%%%%%%%%
\section{Introduction} \label{sec:introduction}

% The bounds of imitation learning
Imitation learning (IL) seeks to adopt optimal policies directly from expert demonstrations.
It mitigates the challenge of explicit reward engineering where poor designs can lead to sub-optimal policies with disastrous behaviors~\cite{amodei_concrete_2016}.
Compared to conventional Reinforcement Learning (RL) algorithms, IL only requires expert examples, which can be much simpler to obtain desired behaviors than providing hand-crafted rewards.
This can be highly appealing in many real-world scenarios where designing a reward function to achieve ideal behaviors requires tremendous effort or is even infeasible in complex systems, {\it e.g.}, autonomous robots \cite{hadfield-menell_cooperative_2016}, trajectory prediction \cite{teranishi_trajectory_2020}, and autonomous driving \cite{bojarski_end_2016}, etc.

Towards robust imitations, recent works on IL have integrated adversarial learning~\cite{ho_generative_2016} which has shown to be highly efficient and effective.
\fix{Rewards are {\it implicitly} inferred from the discriminator, which is trained along with the policy. Therefore, no reward function can be recovered via such methods. This limits IL methods to not generalize outside the environment where it was trained on \cite{fu_learning_2018, yu_meta-inverse_2019}.}
On the other hand, Inverse Reinforcement Learning (IRL) \cite{russell_learning_1998, ng_algorithms_2000} aims to infer a reward function from expert demonstrations to train the agent policy.
The inferred reward functions by IRL can be transferred to different environments.
As reward functions that define the underlying intention of the experts are portable, an agent can be re-optimized in changing dynamics.
Hence, IRL shows high generalization capability to different environments \cite{fu_learning_2018, abbeel_apprenticeship_2004}.

\begin{figure}[t]
    \centering
    \includegraphics[width=0.5\textwidth]{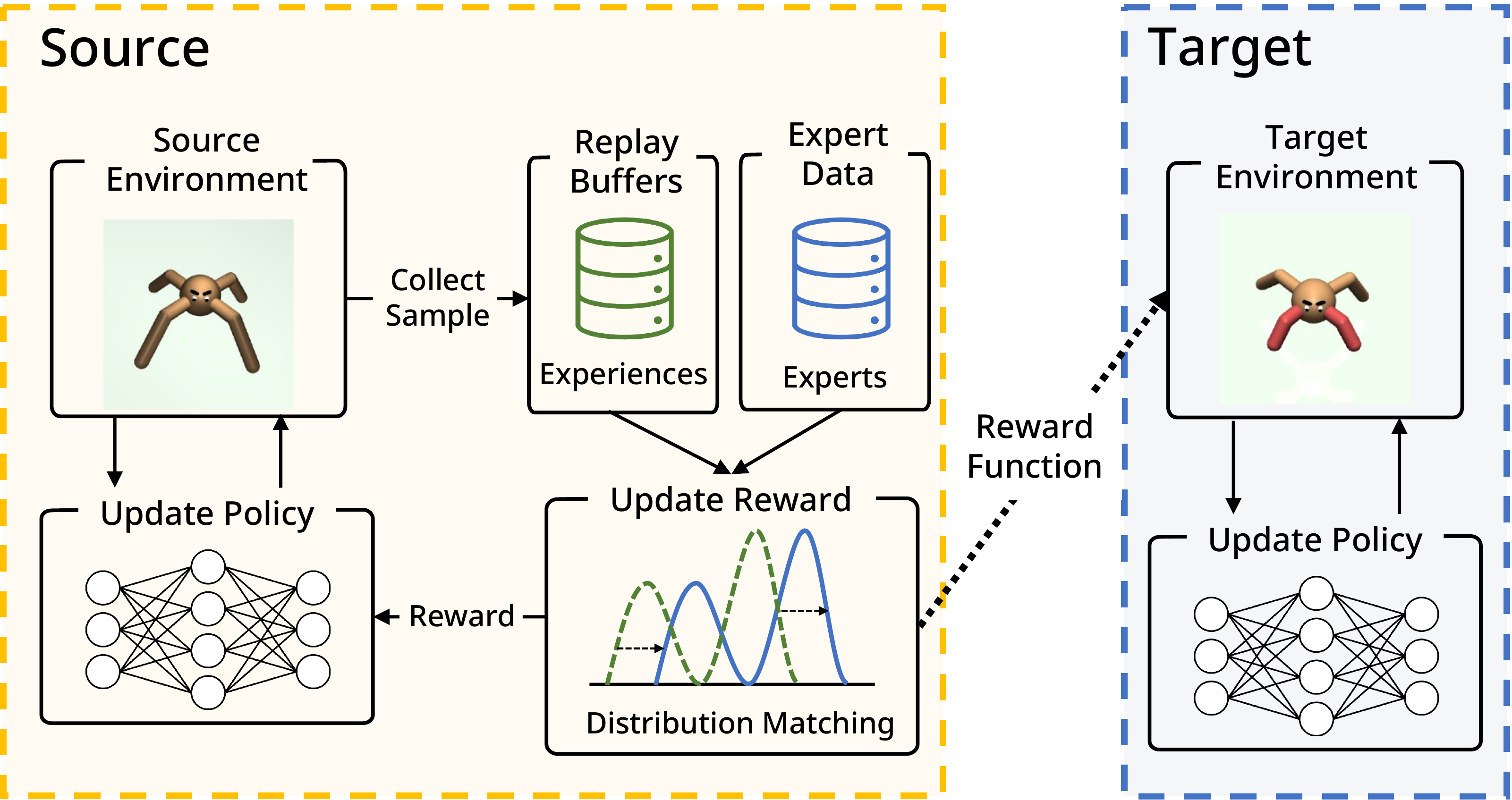}
    \label{fig:opirl}
    \caption{\textbf{Overview of OPIRL framework:} OPIRL (1) achieves high sample efficiency and (2) recovers a robust reward function that can be transferred and generalize across different environments by applying off-policy training in distribution matching.}
    \vspace{-0.3cm}
\end{figure}

% Sample efficiency problem
Although seemingly pleasant and appealing, IRL still has multiple problems, such as ambiguity in reward functions~\cite{ng_algorithms_2000} and sample inefficiency~\cite{kostrikov_discriminator-actor-critic_2018, wu_efficient_2020}.
In our paper, a "sample efficient manner" indicates that an agent requires fewer interactions with the environment, in contrast to some other papers implying the number of expert demonstrations needed to learn optimal reward/policy \cite{yu_meta-inverse_2019}.
In this work, we mainly look into the sample inefficiency of IRL.
Prior arts such as AIRL \cite{fu_learning_2018} require a large amount of expert data and frequent interactions with the environment to recover optimal reward function and policy due to \fix{sampling from the policy that is being trained on ({\it i.e.,} on-policy samples)}.
As on-policy methods rely on Monte Carlo estimations, it suffers from high variance in gradient estimates, which is subsided via intensive sampling \cite{schulman_proximal_2017, blonde_sample-efficient_2019}.
However, exhaustive interactions with the environment can become expensive.
{\it Off-policy} transitions can mitigate the issue by storing previous samples in a replay buffer instead of intensively collecting a large amount of on-policy experience after each policy update.
In the context of IL, several {\it off-policy} algorithms have been proposed \cite{blonde_sample-efficient_2019, sasaki_sample_2018, kostrikov_imitation_2019, zhu_off-policy_2021}.
Especially, OPOLO \cite{zhu_off-policy_2021} removes {\it on-policy} dependencies in distribution matching via equation transformation. 
Nonetheless, no IRL methods have theoretically derived an off-policy IRL algorithm with portable reward functions.

Recent works on IRL shed light on distribution matching \cite{ghasemipour_divergence_2019}, which aims to minimize the difference of stationary distributions between the expert and the learned agent.
Prior approaches use on-policy samples to obtain an accurate distribution of the learning policy.
Inspired by OPOLO, we formulate Off-Policy Inverse Reinforcement Learning (OPIRL) based on the idea of distribution matching and Adversarial Inverse Reinforcement Learning (AIRL).
Furthermore, to encourage active explorations at the beginning stages of training, we adopt mode-covering behavior~\cite{nachum_improving_2016} by integrating behavior cloning loss and Q-Filter~\cite{nair_overcoming_2018}.
OPIRL holds each desired property of prior IL and IRL methods: OPIRL (1) achieves high sample efficiency like recent off-policy IL methods, and (2) generalizes to unseen environments like prior IRL methods.
We demonstrate that OPIRL shows state-of-the-art results on policy performance benchmarks while reducing the interactions with the environment.
Furthermore, we show that OPIRL can generalize to different environments by inferring a transferable reward function, while other IRL methods could not reach the performance of OPIRL, and IL methods fail to generalize. \\
\quad \textbf{Contributions.} Our main contributions are as follows:
\begin{itemize}
    \item We present a novel off-policy IRL method by theoretically deriving a fully off-policy dependent objective function in an IRL setting, which can extract trained reward functions, unlike other prior works. 
    \item We demonstrate that our proposed method outperforms state-of-the-art IRL methods in terms of sample efficiency and shows higher generalizability compared to imitation learning methods through commonly used baselines and robotic tasks.
\end{itemize}
Implementations of OPIRL \fix{are} available at \url{https://github.com/sff1019/opirl}.

% Related Works %%%%%%%%%%%%%%%%%%%%%%%%%%%%%%%%%%%%%%%%%%%%%%%%%%%%%%%%%%%%%%%%%%%%%%%%%%%%%%%%
\section{Related Work} \label{sec:related_work}

\textbf{Inverse reinforcement learning} is a problem setting to learn a reward function from a set of expert trajectories \cite{ng_algorithms_2000}.
MaxEntIRL recovers reward functions by minimizing the forward KL divergence in trajectory space under the maximum entropy RL framework \cite{ziebart_maximum_nodate}.
To retrieve policy while learning reward functions directly, recent works combine generative adversarial network (GAN) \cite{goodfellow_generative_2014} to training.
GAN-GCL \cite{finn_guided_2016} uses GAN to optimize an MLE objective over trajectories.
AIRL \cite{fu_learning_2018} improves upon this to recover a reward function while simultaneously learning the policy.
Instead of using adversarial learning, \textit{f}-IRL~\cite{ni_f-irl_2020} solves the state marginal matching problem to infer the reward function.
Although such works show prominent improvement in robustness and performance, many remain data-hungry due to the on-policy data transitions.

\textbf{Imitation learning} does not recover reward functions; instead, it tries to acquire optimal policy from demonstrations provided by an expert policy.
GAIL \cite{ho_generative_2016} uses the GAN formulation in training, allowing it to be more sample efficient than behavior cloning (BC) in terms of the number of expert demonstrations.
However, similar to IRL methods, it suffers sample inefficiency.
To mitigate this problem, several \textit{off-policy} IL methods have recently been proposed.
Sample-efficient Adversarial Mimic (SAM) \cite{blonde_sample-efficient_2019} uses off-policy actor-critic to remove the on-policy dependencies.
Sasaki et al. \cite{sasaki_sample_2018} propose an algorithm that incorporates an off-policy actor-critic algorithm to optimize the policy.
Similarly, Discriminator-Actor-Critic (DAC) \cite{kostrikov_discriminator-actor-critic_2018}, an extension of GAIL, uses observations stored in the replay buffer instead of on-policy transitions for its policy updates.
Nonetheless, although DAC shows empirically better results in terms of sample efficiency, it deviates from its theoretically correct objective because it ignores computing the importance sampling term.
To remove such discrepancy, several off-policy IL methods utilize distribution matching algorithms: ValueDICE \cite{kostrikov_imitation_2019} and OPOLO \cite{zhu_off-policy_2021}.
ValueDICE incorporates distribution correction estimation~\cite{nachum_dualdice_2019} to remove on-policy dependency.
However, the policy objective contains logarithms and exponential expectations, which introduce biases in its gradients \cite{sun_softdice_2021}.
On the other hand, OPOLO adopts an off-policy transition to IL in principle manner by deriving an upper bound of the IL objective, which removes such biases.

Our work is built on the findings of OPOLO to improve sample efficiency in inverse reinforcement learning framework while recovering reward functions to achieve high generalization across different environments, which imitation learning methods ({\it e.g.,} DAC, OPOLO, etc.) are prone to fail.

% Preliminaries %%%%%%%%%%%%%%%%%%%%%%%%%%%%%%%%%%%%%%%%%%%%%%%%%%%%%%%%%%%%%%%%%%%%%%%%%%%%%%%%
\section{Preliminaries}\label{sec:preliminaries}

We consider a Markov Decision Process (MDP), defined by the tuple $(\mathcal{S}, \mathcal{A}, \mathcal{P}, r, \rho_0, T)$, where $\mathcal{S}$ and $\mathcal{A}$ denote the state and action space, \fix{transition dynamics $\mathcal{P}: \mathcal{S} \times \mathcal{A} \times \mathcal{S} \rightarrow [0, 1]$}, $r(s,a)$ as the reward function, initial state-action distribution $\rho_0(s, a)$, and horizon $T$. 
The goal of {\it forward} reinforcement learning is to find the optimal policy $\pi^\ast$ that maximizes the expected entropy-regularized discounted reward \cite{ziebart_maximum_nodate}:
\begin{equation} \label{eq:maxent}
    \pi^\ast = \argmax_\pi \mathbb{E}_{\rho_{\pi}} \left[ \sum_{t=0} ^{T} \gamma^t \left( r(s_t, a_t) + \alpha \mathcal{H} (\pi(\cdot  | s_t)) \right) \right]
\end{equation}
where $\rho_\pi$ is the state-action distribution, and $\alpha > 0$ is the entropy temperature.

\subsection{Off-Policy Learning from Observation}
Our approach takes its inspiration from OPOLO~\cite{zhu_off-policy_2021} which, similar to prior off-policy RL algorithms, removes all on-policy transitions and incorporates replay buffers to improve sample efficiency.
The objective function of various IL/IRL methods can be written as minimization of statistical divergence \cite{ghasemipour_divergence_2019}.
Using this notion, \citet{zhu_off-policy_2021} sets the objective function using state distribution as
\begin{equation} \label{eq:opolo_objective}
    \min J(\pi) = \mathrm{KL} \left( \distagent(s, s') \| \distexp(s, s') \right).
\end{equation}
\citeauthor{zhu_off-policy_2021} first integrates state-action distribution of the replay buffer $\distrb(s, a)$ by creating an upper bound.
They further introduce an upper-bound of KL-divergence for stable learning \cite{sun_softdice_2021} by showing that when $f(x) = \frac{1}{p} |x|^p$, KL-divergence is upper-bounded by \fdivergence:
\begin{equation} \label{eq:kl_upper}
    \mathrm{KL} (P \| Q) \leq \mathbb{D}_{f} (P \| Q).
\end{equation}
The introduction of \fdivergence can become challenging.
Therefore, to make the objective function more approachable, \citeauthor{zhu_off-policy_2021} transforms \fdivergence to its variational  \cite{nguyen_estimating_2010} form using dual function:
\begin{equation} \label{eq:f_divergence_variational}
\begin{split}
    - &\mathbb{D}_f [\distagent (s, a) \| \distrb (s, a) ] \\
     &= \inf_{x: \state \times \action \rightarrow R} \mathbb{E}_{\distagent} [-x (s, a)] + \mathbb{E}_{\distrb} [f_\ast (x (s, a))]           
    \end{split}
\end{equation}
where $f_\ast(\cdot)$ is the convex conjugate of $f(\cdot)$.
In our case, the convex conjugate function of $f(x) = \frac{1}{p} |x|^p$ is:
\begin{equation} \label{eq:convex_conjugate}
    f_\ast(x_\ast) = \frac{1}{q} |x_\ast|^q, \ 1 \leq q \leq \infty, \ \mathrm{where} \  \frac{1}{p} + \frac{1}{q} = 1.
\end{equation}
Using this transformation, the objective function can be re-written with expectations of the state distribution of the current policy and the replay-buffer:
\begin{align}
    \max_\pi \min_x J_{opolo} (\pi, x) := \mathbb{E}_{\distagent}& \left[ \log \frac{\distexp(s, s')}{\distrb (s, s')} - x(s, a) \right] \nonumber \\ 
    &+ \mathbb{E}_{\distrb} \left[ f_\ast (x(s, a)) \right]. 
    \end{align}
\citeauthor{zhu_off-policy_2021} further show that $\log \frac{\distexp (s, s')}{\distrb (s, s')} - x(s, a)$ can be interpreted as a {\it synthetic} reward.
Using this interpretation, we can perform the following change of variables using Q-function:
\begin{equation} \label{eq:change_of_variables} 
    x(s, a) = (\mathcal{B}^\pi Q_\phi - Q_\phi) (s, a).
\end{equation}
Applying this change of variables and some telescoping, they derived an objective function that solely depends on the initial state distribution and off-policy distribution.
This change of variables technique has been widely used in recent works \cite{nachum_dualdice_2019, kostrikov_imitation_2019}.
\citeauthor{zhu_off-policy_2021} makes use of GAN training to estimate the log term $\frac{\distexp (s, s')}{\distrb (s, s')}$.

\subsection{Inverse Reinforcement Learning}
{\it Inverse} reinforcement learning seeks to infer the reward function $r(s, a)$ from a given set of demonstrations $\mathcal{D} = \{\tau_1, \dots, \tau_N\}$ from an expert policy $\pi_\mathrm{exp}$, where $\tau$ is the trajectory.
MaxEntIRL \cite{ziebart_maximum_nodate}, a method built on the maximum entropy RL framework, can be interpreted as solving the maximum likelihood problem:
\begin{equation}
    \max_\theta J(\theta) = \max_\theta \mathbb{E}_{\tau \sim \mathcal{D}} [\log \rho^{\pi_\theta} (\tau)]
\end{equation}
where $p_\theta (\tau) \propto p(s_0) \sum_{t=0}^{T} p(s_{t+1} | s_t, a_t) \exp (r (s, a) / \alpha)$. 
Adversarial Inverse Reinforcement Learning (AIRL) \cite{fu_learning_2018} involves GAN formulation to acquire solutions for MaxEntIRL.
The discriminator in AIRL is structured as
\begin{equation} \label{eq:airl_discriminator}
    D_{\omega, \Phi}(s, a) = \frac{\exp (h_{\omega, \Phi}(s, a))}{\exp (h_{\omega, \Phi}(s, a) + \pi_\theta (a | s))}
\end{equation}
where $h_{\omega, \Phi}(s, a) = r_\omega(s, a) + \gamma g_\Phi (s') - g_\Phi (s)$, reward function $r_\omega(s, a)$, and \fix{reward shaping term} $g_\Phi(s)$ and $g_\Phi(s')$.
\fix{With some assumptions, it has been proven that $r_\omega(s, a)$ and $g_\Phi(s)$ will recover ground-truth reward and optimal value function up to a constant \cite{fu_learning_2018}.}
AIRL policy optimization can be interpreted as solving the MaxEntIRL problem by minimizing the reverse KL-Divergence, $\min \mathrm{KL}(\distagent (s, a) || \distexp (s, a))$, where $\distagent$ and $\distexp$ are the state-action distribution of the current agent and the expert respectively \cite{ghasemipour_divergence_2019}.
% %%%%%%%%%%%%%%%%%%%%%%%%%%%%%%%%%%%%%%%%%%%%%%%%%%%%%%%%%%%%%%%%%%%%%%%%%%%%%%%%
\section{OPIRL: Off-Policy Inverse Reinforcement Learning via Distribution Matching} \label{sec:method}
In this section, we will introduce our algorithm: OPIRL.
\citeauthor{zhu_off-policy_2021} shows the objective function of OPOLO as the discrepancy between the agent and expert's state-transition distribution.
On the contrary, the objective of IRL is to model an agent taking actions in a given environment.
Therefore, we first show how we incorporate off-policy learning in \fix{IRL} setting, and then show how we further improve the efficiency to train OPIRL.
\subsection{Off-Policy in Objective Function} \label{subsec:opirl}
As stated in Sec.~\ref{sec:preliminaries}, AIRL minimizes the divergence between agent and expert state-action distribution.
This training process can become sample-inefficient as computing the state-action distribution of the agent requires \textit{on-policy} interactions with the environment.
To resolve this issue, we will adopt {\it off-policy} distribution $\distrb (s, a)$ into our objective function.
The reverse KL-Divergence between agent and expert state-action distribution can be rewritten using state-action distribution of the replay-buffer $\distrb$:
\begin{align} \label{eq:opirl_with_rb}
     &\mathrm{KL} (\distagent (s, a) \| \distexp (s, a)) \nonumber \\
    &= \mathbb{E}_{\distagent} \left[ \log \frac{\distrb (s, a)}{\distexp (s, a)} \right] + \mathrm{KL} (\distagent (s, a) \| \distrb (s, a))       
\end{align}
 where the replay buffer \cite{lin_self-improving_1992} is a structure used in many off-policy learning methods in which past experiences are re-used to improve sample efficiency and performance.
 \fix{Unlike OPOLO, we do not have to create an upper bound to incorporate $\distrb$, which can lead to more optimal performance.}
 
 \fix{
 For stable training, we follow prior works via setting the upper bound using Eq~(\ref{eq:kl_upper}):
 }
\begin{equation} \label{eq:opirl_with_f}
\min_\pi J(\theta) := \mathbb{E}_{\distagent} \left[ \log \frac{\distrb (s, a)}{\distexp (s, a)} \right] + \mathbb{D}_{f} [\distagent (s, a) \| \distrb (s, a)].        
\end{equation}
\fix{As mentioned in Sec. \ref{sec:preliminaries}, the introduction of {\it f}-Divergence can be complicating, thus we remove this by using Eq~(\ref{eq:f_divergence_variational}):}
\begin{align}
    &\min J(\theta) \nonumber \\
    &:= \max_\pi \mathbb{E}_{\distagent} \left[ - \log \frac{\distrb (s, a)}{\distexp (s, a)} \right] - \mathbb{D}_f [\distagent (s, a) \| \distrb (s, a)] \nonumber \\
    &= \max_\pi \mathbb{E}_{\distagent} \left[\log \frac{\distexp (s, a)}{\distrb (s, a)} - x(s, a) \right] - \mathbb{E}_{\distrb} [f_\ast (x(s,a))] \label{eq:opirl_after_dual}.
\end{align}
\fix{To enable off-policy learning, it is required to remove all on-policy interactions $\distagent$ from the objective function. 
We take inspiration from the change of variables Eq.~(\ref{eq:change_of_variables}):
}
\begin{equation} \label{eq:objective_fn}
\begin{split}
& \max_\pi \min_{x: \mathcal{S} \times \mathcal{A} \rightarrow R} J(\pi, x) = \max_\pi \min_{Q: \state \times \action \rightarrow R} J(\pi, Q_\phi) \\
&= \mathbb{E}_{\distagent}\left[\log \frac{\distexp \left(s, a\right)}{\distrb \left(s, a\right)}-\left(\mathcal{B}^{\pi} Q_\phi-Q_\phi\right)(s, a)\right] \\
& \quad +\mathbb{E}_{\distrb}\left[f_{*}\left(\left(\mathcal{B}^{\pi} Q_\phi-Q_\phi\right)(s, a)\right)\right] \\
&= (1 - \gamma) \mathbb{E}_{s_0 , a_0} [Q(s_0, a_0)] + \mathbb{E}_{\distrb}[f_\ast ((\mathcal{B}^\pi Q_\phi - Q_\phi)(s, a))].
\end{split}
\end{equation}
This final objective function is fully \textit{off-policy}, thus can be trained by using off-policy transitions sampled from replay buffer $\distrb$.
We can estimate the log term $\log  \frac{\distexp (s, a)}{\distrb (s, a)}$ in $\mathcal{B}^\pi Q$ \fix{through adversarial learning.}

\subsection{Learning Rewards via Inverse Reinforcement Learning} \label{subsec:opirl_disc}
\label{sec:method_reward}

We use the same discriminator structure as AIRL (Eq.~(\ref{eq:airl_discriminator})).
Through the training of this discriminator, we can obtain an explicit reward function, unlike prior works \cite{zhu_off-policy_2021, kostrikov_imitation_2019}.
We modify the objective function of the discriminator to depend on the state-action distribution of the expert and the replay buffer:
\begin{equation}
\begin{split}
    L_{\omega, \Phi} (s, a) = \sum_{t=1} ^{T} - & \mathbb{E}_{\distexp} \left[ \log D_{\omega, \Phi} (s_t, a_t) \right] \\
    &- \mathbb{E}_{\distrb} \left[ \log(1 - D_{\omega, \Phi} (s_t, a_t) \right],
\end{split}
\end{equation}
\fix{and when reaching optimality $\log D_{\omega, \Phi}^{\ast} (s, a) - \log(1 - D_{\omega, \Phi}^{\ast} (s, a))= \log  \frac{\distexp (s, a)}{\distrb (s, a)}$ \cite{goodfellow_generative_2014}.
Therefore, in our setting the log term can be considered as the reward}:
\begin{equation} \label{eq:airl_original_reward}
    \begin{split}
     \tilde{r}_{\omega, \Phi}(s, a) =& \ \log D_{\omega, \Phi}(s, a) - \log(1 - D_{\omega, \Phi}(s, a)) \\
            =& \ h_{\omega, \Phi}(s,a) - \log \pi_\theta (a | s),
    \end{split}
\end{equation}
which can then be interpreted as an entropy regularized reward function.
Following empirical findings  \cite{yu_multi-agent_2019, arnob_off-policy_2020}, we remove the entropy regularized term and reward shaping term $g_\Phi(\cdot)$ in Eq.~(\ref{eq:airl_original_reward}) and use the form $\tilde{r}_{\omega, \Phi}(s, a) = r_{\omega}(s, a)$.

\subsection{Modification for Efficient Training}\label{sec:efficient_training}
In order to improve the efficiency of the training, we add two modifications to the objective function Eq.~(\ref{eq:objective_fn}).
First, we modify the policy updates.
It is widely known that adding an entropy regularization in policy updates improves performance in continuous control tasks \cite{haarnoja_reinforcement_2017, nachum_algaedice_2019}. 
Therefore, we add a causal entropy term $\log \pi (a | s)$ in Eq.~(\ref{eq:objective_fn}):
\begin{equation}
    \begin{split}
    J(\pi_\theta, Q_\phi) =& (1 - \gamma) \mathbb{E}_{s\sim s_0} [Q_\phi (s, \pi_\theta (s)] \\
    &+ \mathbb{E}_{(s, a) \sim \distrb (s, a)} [f_\ast (\delta (s, a, r, s', a')) ],
    \end{split}
\end{equation}
where $s'$ and $a'$ are the next state and action sampled from the policy, and 
\begin{equation} \label{eq:causal_entropy}
    \begin{split}
        \delta(s, a, r, s', a') =& \ \tilde{r}_{\omega, \Phi}(s, a) - \eta \log \pi_\theta(a' | s')\\
         &+ \gamma Q_\phi (s', a') - Q_\phi (s, a),
    \end{split}
\end{equation}
where $\eta$, also known as the temperature variable in \cite{haarnoja_soft_2018}, is learned during training.

Secondly, we adopt mode-covering behaviors in our policy updates \cite{zhu_off-policy_2021, nachum_improving_2016, wu_squirl_2020}.
Ghasemipour \textit{et al.} \cite{ghasemipour_divergence_2019} hypothesized that a mode-seeking behavior, which can be seen in reverse KL-Divergence such as ours, is desirable in RL scenarios as it cares more about the trajectories where the expert has visited.
However, such behavior is less likely to explore regions expert has not covered.
It tends to fall into local optimum, resulting in sub-optimal performance \cite{nachum_improving_2016}.
On the other hand, mode-covering enables policies to explore more widely. 
This behavior can be achieved via forward KL-Divergence, such as behavior cloning.
Hence, we incorporate Behavior Cloning Loss and Q-Filter \cite{nair_overcoming_2018, ota_trajectory_2020} into our policy updates to encourage active exploration at the beginning stages of training while avoiding sub-optimal actions being chosen.
\begin{eqnarray}
    L_{BC} = 
    \begin{cases}
        \left( \pi(s_i) - a_i \right)^{2},& \text{if }  Q_\phi(s_i, \pi (s_i) ) \leq Q_\phi(s_i, a_i) \\
        0, & \text{otherwise}.
    \end{cases}
\end{eqnarray}
We show the overview of our algorithm in Algorithm~\ref{algo:opirl}.

\begin{algorithm}
\setstretch{1.0}
\SetAlgoLined
\textbf{Input:} A set of expert trajectories $\tau_i ^E$\\
\fix{Initialize $R$ \Comment{\textbf{Replay buffer for IRL}}}\\
\For{\fix{each iteration}}{
    \fix{
    \For{each environment step}{
    Sample action $a_t \sim \pi_\theta (\cdot | s_t)$ \\
    Sample next state $s_{t+1} \sim p(\cdot | s_t, a_t)$ \\
    Obtain reward $r_t \leftarrow \tilde{r}_{\omega, \Phi} (s_t, a_t)$ \\
    Update buffer $R \leftarrow R \cup \{(s_t, a_t, r_t, s_{t+1}) \}$ \\
    }
   
    \For{each IRL gradient step}{
    Sample $s$ and $a$ from replay buffer $R$ \\
     Update reward function $\omega \leftarrow \omega - \alpha \nabla_\omega L_{\omega, \Phi}$ \\
     Update value function $\Phi \leftarrow \Phi - \alpha \nabla_\Phi L_{\omega, \Phi}$ \\
     }
    \For{each RL gradient step}{
    Sample $s$ and $a$ from replay buffer $R$ \\
     \eIf{$Q(s, a) \leq Q(s, \pi_\theta(s))$}{
        $L_{BC} (\pi_\theta) = \left(\pi_\theta (s) - a \right)^2$
        }{
        $L_{BC} (\pi_\theta)$ = 0
        }   
        Update policy $\theta \leftarrow \theta - \beta (\lambda_1 \nabla_\theta J + \lambda_2 \nabla_\theta L_{BC})$ \\
        Update critic $\phi \leftarrow \phi - \beta \nabla_\theta J$
    }
    }
 }
 \caption{OPIRL: Off-Policy IRL}
 \label{algo:opirl}
\end{algorithm}

% %%%%%%%%%%%%%%%%%%%%%%%%%%%%%%%%%%%%%%%%%%%%%%%%%%%%%%%%%%%%%%%%%%%%%%%%%%%%%%%%
\section{Experiments} \label{sec:experiments}

\begin{figure*}[t]
    \centering
    \includegraphics[width=\linewidth]{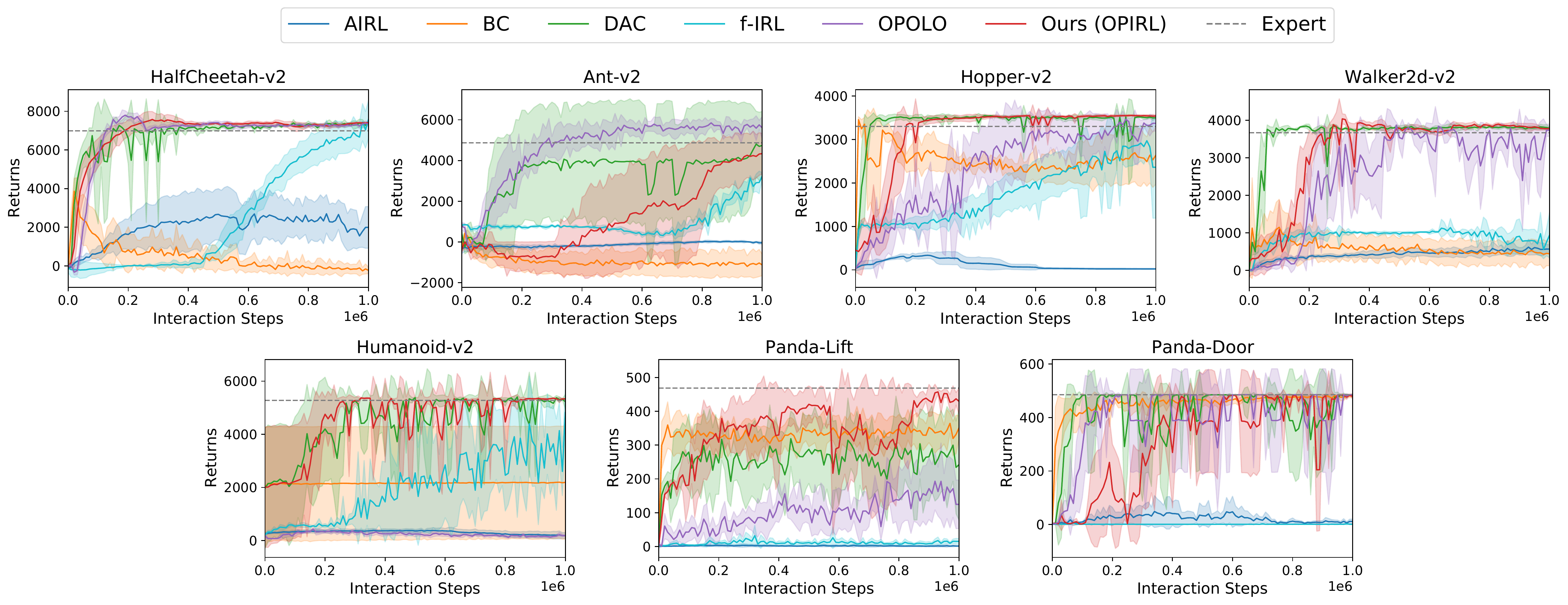}
    \caption{Training curves of different algorithms. Locomotion tasks use {\it single} expert trajectory and robotic tasks use sixteen trajectories. The expert policy performance is shown in a gray horizontal line, and we run each experiment independently with 5 seeds and plot the average and $\pm 1$ standard deviation with solid lines and shaded regions, respectively. OPIRL outperforms other IRL methods while achieving comparable performance with off-policy IL methods.}
    \label{fig:policy_performance}
    \vspace{-4mm}
\end{figure*}

To evaluate the performance of our proposed algorithm, we conduct experiments to address the following questions:
\begin{enumerate}
    \item \textbf{Policy Performance}: Can OPIRL recover the expert policy on imitation learning tasks with a smaller number of interactions with the environment?
    \item \textbf{Reward Robustness}: Can OPIRL learn a robust reward function that can be transferred to environments that have different structures or dynamics?
\end{enumerate}

To answer these questions, we compare our algorithm against IL methods that directly learn the policy: Behavior Cloning (BC), OPOLO \cite{zhu_off-policy_2021} and DAC \cite{kostrikov_discriminator-actor-critic_2018}, and IRL methods that learn the reward function and the policy: \textit{f}-IRL \cite{ni_f-irl_2020}, and AIRL \cite{fu_learning_2018}.
Note that OPOLO and DAC are off-policy IL methods.
The first question is evaluated on 5 commonly used MuJoCo locomotion tasks \cite{todorov_mujoco_2012}, and 2 robotic tasks using Robosuite \cite{zhu_robosuite_2020}.
The second question is tested by transferring the trained rewards/policy from source to target environment on MuJoCo simulation.
For each task, we train an agent using Soft-Actor-Critic (SAC) \cite{haarnoja_reinforcement_2017} as the expert policy, which is then used to collect a set of expert trajectories.
We run all experiments on 5 seeds for a fair comparison.

\subsection{Policy Performance} \label{subsec:experiments_policy}

To evaluate whether OPIRL can acquire a policy that exhibits high performance like the expert policy in a sample efficient manner, we compare OPIRL with other IL / IRL methods.
We conduct this experiment using a {\it single} expert trajectory as similarly done in prior works \cite{kostrikov_discriminator-actor-critic_2018, ni_f-irl_2020} for locomotion tasks and sixteen for robotic tasks.
Robotic tasks require more expert as only a single sequence of expert is provided per sample, whereas locomotion tasks have multiple sequences in a single expert demonstration.

We show the learning curves of the average return using one expert trajectory in Fig.~\ref{fig:policy_performance}.
It shows that OPIRL successfully recovers expert performance throughout the seven tasks.
Compared to OPIRL, {\it f}-IRL requires much more steps to converge across all tasks.
We can also see that AIRL seems to fail on all tasks, similar to the findings of \cite{ni_f-irl_2020}.

Next, we compare OPIRL to IL methods.
BC also fails to recover a policy, which is mainly due to a lack of training data as it is known to have a covariate-shift problem, hence requiring multiple trajectories for robust learning \cite{blonde_lipschitzness_2021}.
Off-policy based IL methods, DAC and OPOLO, succeeded in recovering expert policy in most tasks.
Overall, we can conclude that OPIRL can learn much more sample efficient than prior IRL methods such as AIRL and \textit{f}-IRL, and perform comparably with off-policy IL algorithms.
Furthermore, OPIRL succeeded in obtaining such performance with a single expert trajectory.

\subsection{Reward Robustness: Transfer Learning} \label{subsec:experiments_transfer}
\begin{figure} [htbp]
     \begin{minipage}[b]{0.49\linewidth}
        \captionsetup{justification=centering}
        \centering
        \subfloat[PointMaze-Left\\ (source)]{\label{fig:3_a}\includegraphics[width=0.6\linewidth]{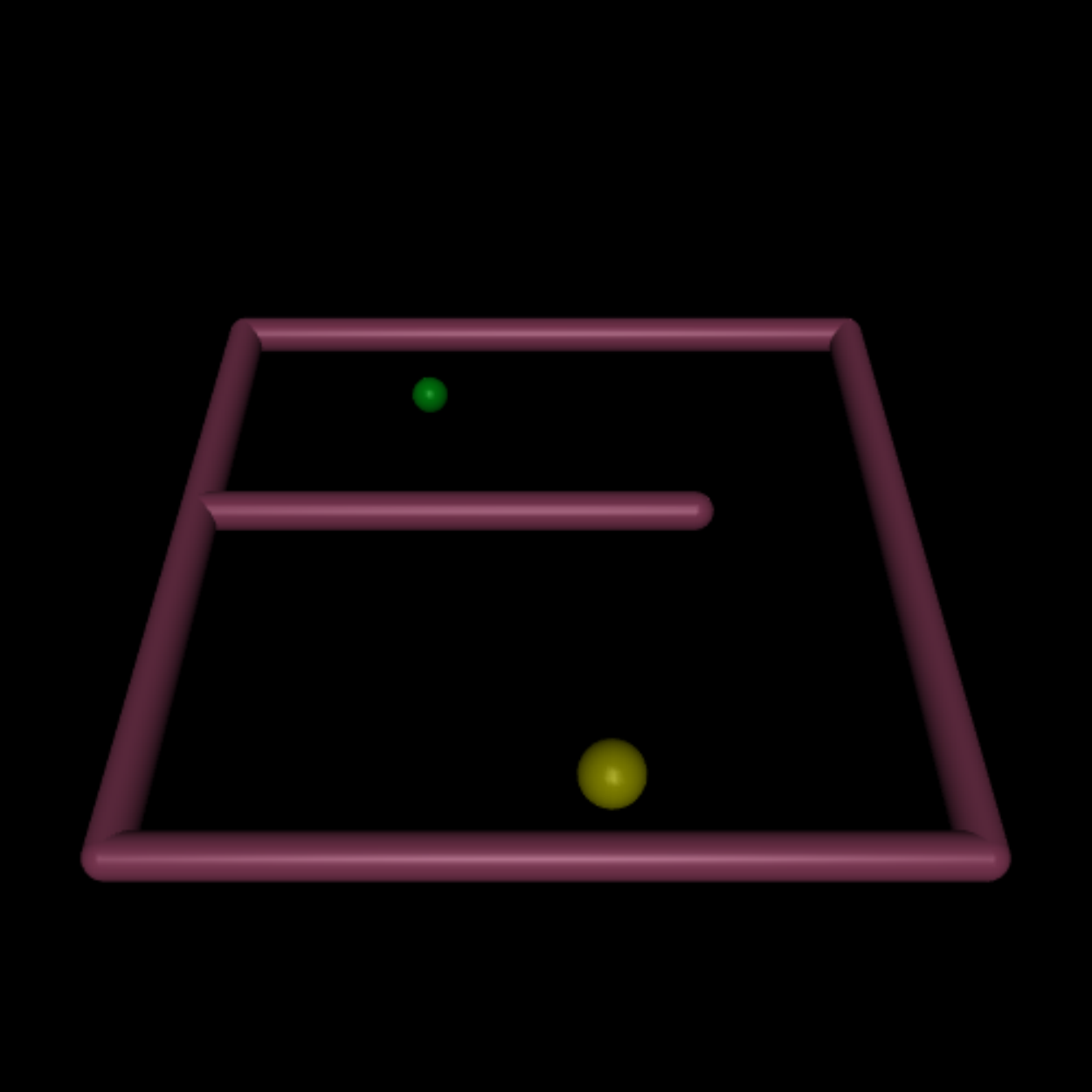}}
        \vspace{0.1cm}
     \end{minipage}
     \begin{minipage}[b]{0.49\linewidth}
        \captionsetup{justification=centering}
         \centering
        \subfloat[PointMaze-Right\\ (target)]{\label{fig:3_b}\includegraphics[width=0.6\linewidth]{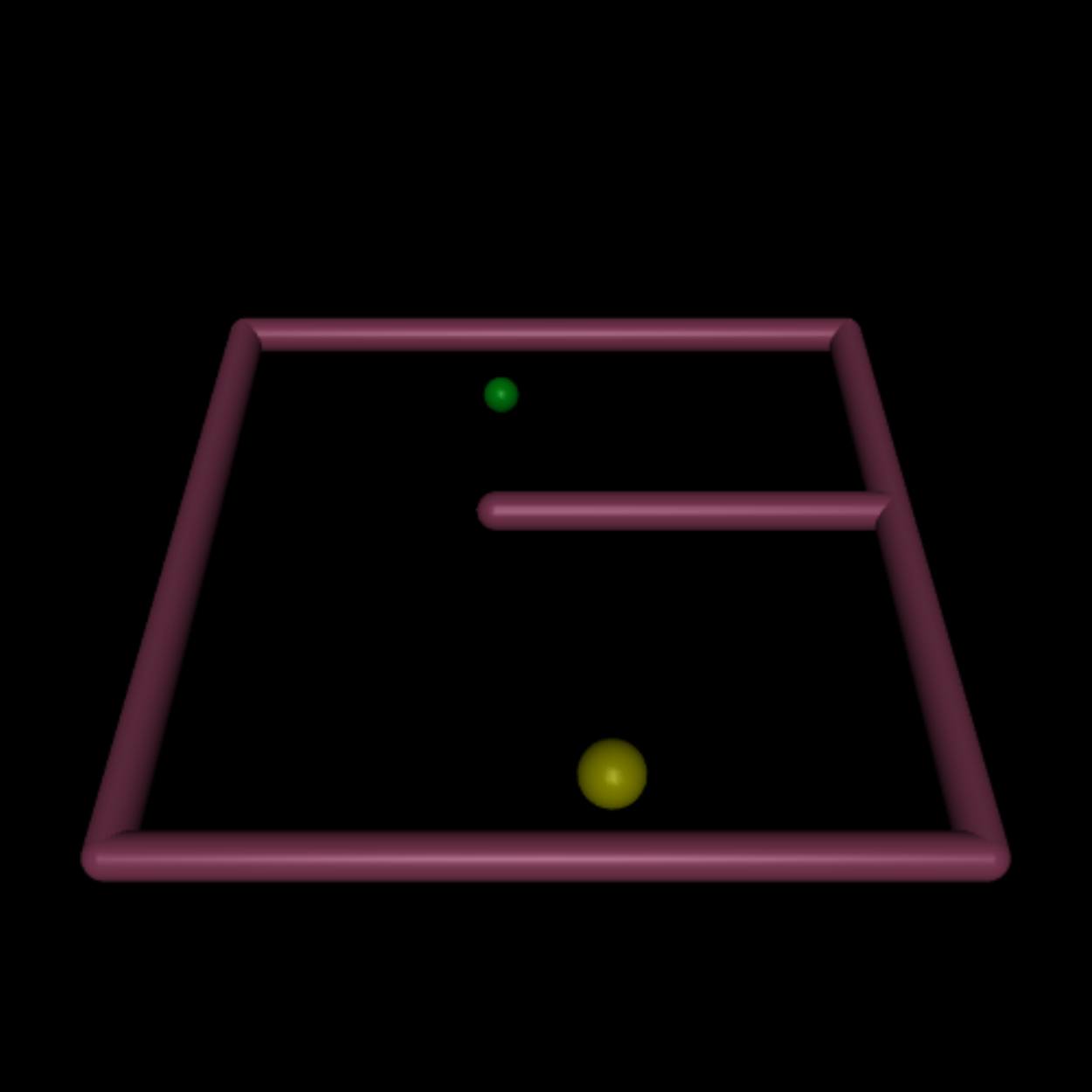}}
        \vspace{0.1cm}
     \end{minipage}
    \begin{minipage}[b]{0.32\linewidth}
        \captionsetup{justification=centering}
        \centering
        \subfloat[CustomAnt\\ (source)]{\label{fig:3_c}\includegraphics[width=0.8\linewidth]{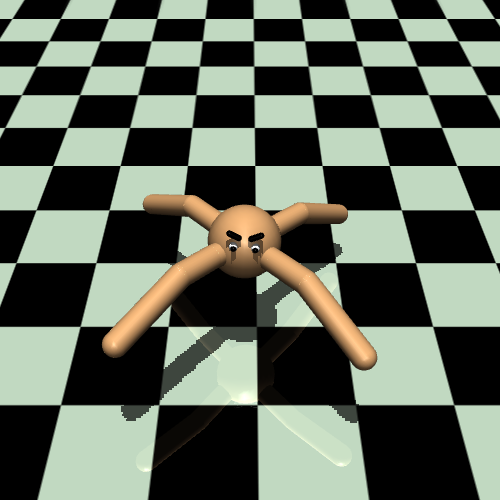}}
    \end{minipage}  
    \begin{minipage}[b]{0.32\linewidth}
        \captionsetup{justification=centering}
        \centering
        \subfloat[BigAnt\\ (target)]{\label{fig:3_d}\includegraphics[width=0.8\linewidth]{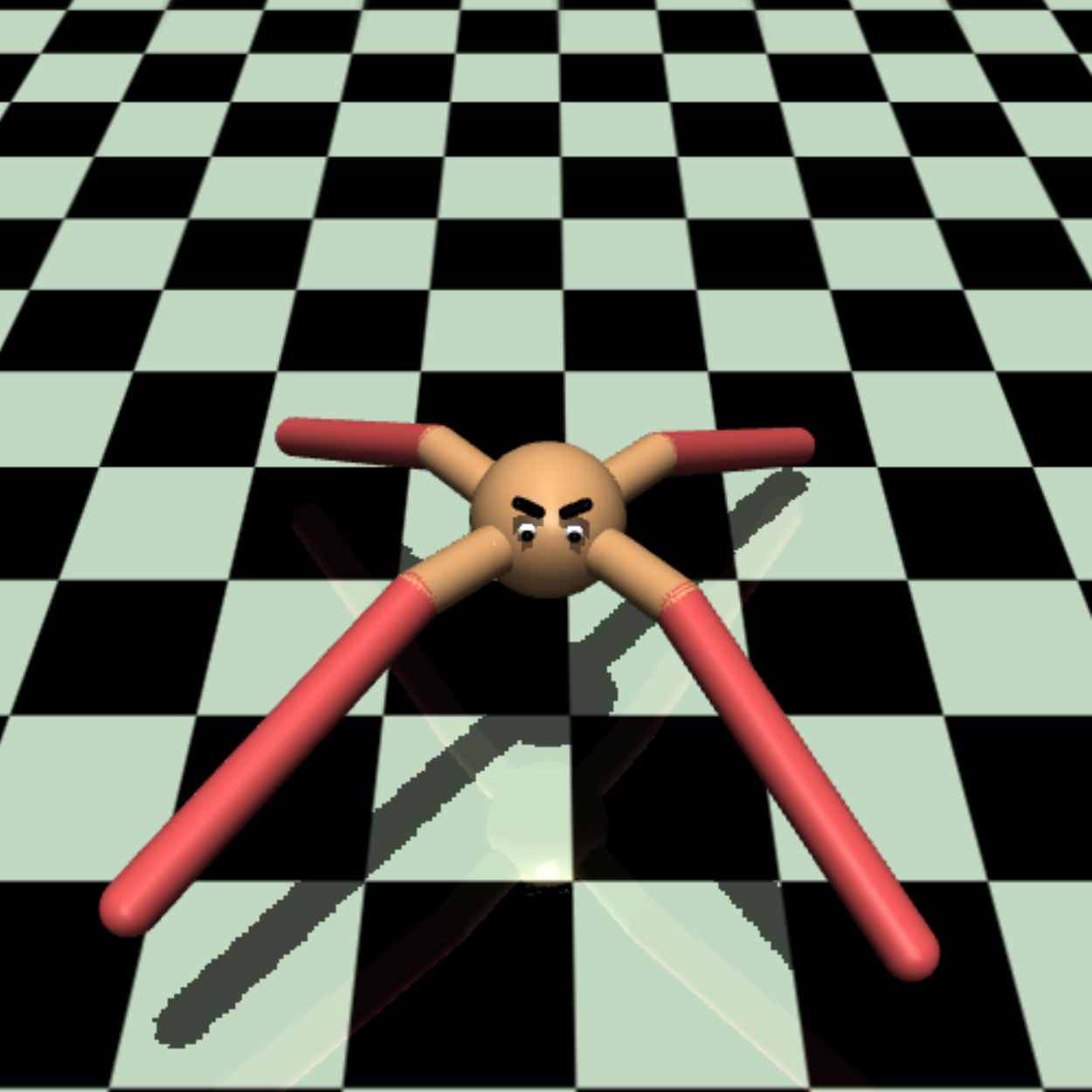}}
    \end{minipage}   
    \begin{minipage}[b]{0.32\linewidth}
        \captionsetup{justification=centering}
        \centering
        \subfloat[AmputatedAnt (target)]{\label{fig:3_e}\includegraphics[width=0.8\linewidth]{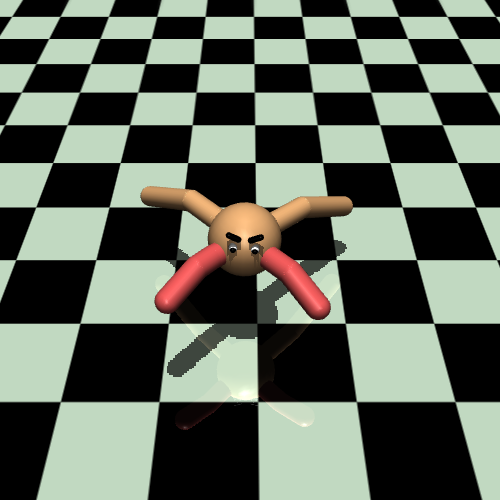}}
    \end{minipage}    
    \caption{Environments for reward robustness experiment. Agents are trained on source environment (leftmost) and evaluated on the target environment (middle or right).}
    \label{fig:transfer_laerning_environments}
\end{figure}
\begin{figure*}[tb]
    \centering
    \includegraphics[width=\textwidth]{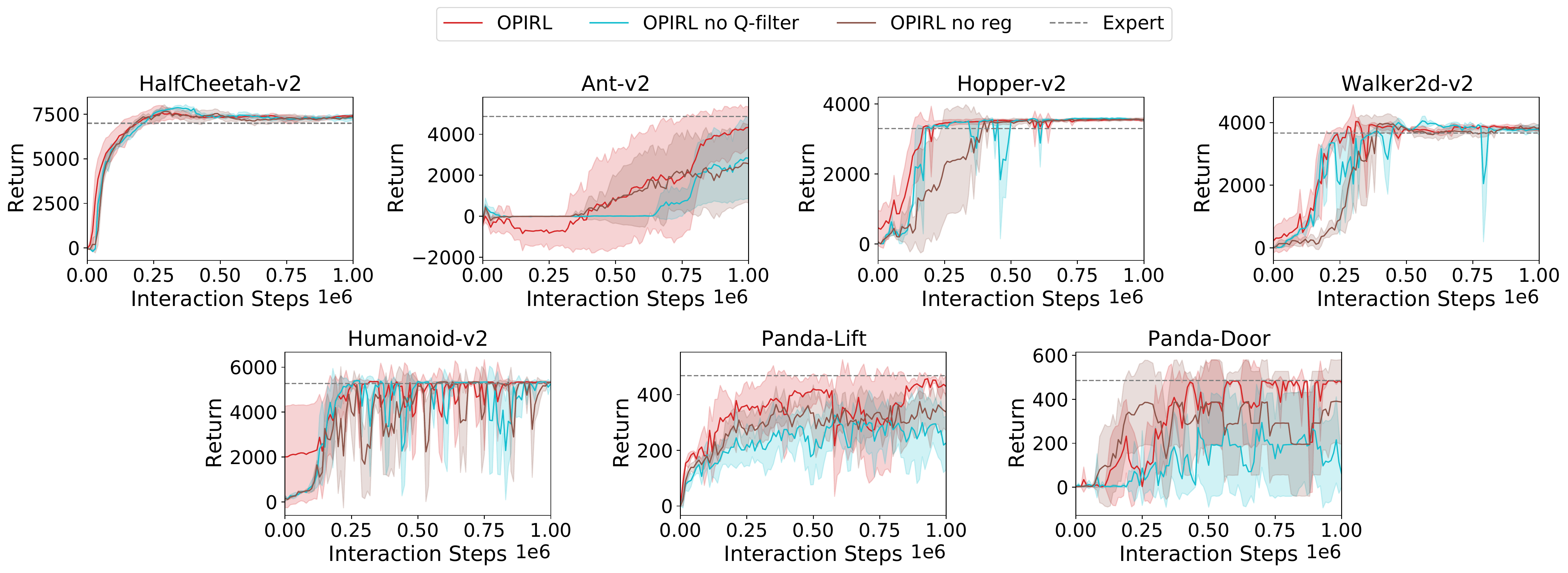}
    \caption{Learning curves of OPIRL with and without regularization. Applying BC Loss+Q-Filter enables the agent to explore more widely at the beginning stages of training, thus converging faster than the agent without such regularization.}
    \label{fig:sample_efficiency}
\end{figure*}

To evaluate if OPIRL can learn a reward function that can generalize to different dynamics, we follow the setups of the AIRL paper~\cite{fu_learning_2018}; we use the learned reward function from the source environment to re-train a policy on the target environment.
As IL methods do not explicitly recover reward functions, we instead transfer the trained policy to the source environment to evaluate it on the target environment.
We conducted experiments on two settings: 1) changing the environment structure, and 2) changing the dynamics of the environment (see Fig.~\ref{fig:transfer_laerning_environments}).

\begin{table}
\caption{Comparison between the final average return after transferring the reward function from source environments to target environments. Each method is evaluated using 20 episodes under 5 random seeds. $^\dag$ methods indicate results on direct policy generalization, and the bold number indicates the best score among each algorithm.}
\label{tab:transfer_learning}
\centering
\resizebox{0.49\textwidth}{!}{\begin{tabular}{lrrrr}
    \toprule
     Method & PointMaze-Right & BigAnt & AmputatedAnt \\ \midrule
     DAC$^{\dag}$ & \fix{$-32.1 (\pm 2.9)$} & \fix{$-4.1 (\pm 25.0)$} & \fix{$-39.4 (\pm 18.9)$}\\
     OPOLO$^{\dag}$ & \fix{$-36.0 (\pm 0.5)$}& \fix{$-23.5 (\pm 13.4)$} & \fix{$-47.2 (\pm 8.3)$} \\
     AIRL & \fix{$-44.4 (\pm 12.0)$} & \fix{$-42.8 (\pm 32.3)$} & \fix{$-9.1 (\pm 33.0)$} \\
     \textit{f}-IRL & \fix{$-120.0 (\pm 31.0)$} & \fix{$554.4 (\pm 317.0)$} & \fix{$463.1 (\pm 113.6)$}\\
     OPIRL & \fix{$\mathbf{-14.4 (\pm 1.2)}$} & \fix{$\mathbf{2024.1 (\pm 1434.5)}$} & \fix{$\mathbf{621.9 (\pm 5.0)}$}  \\
     \midrule
     Ground-truth &$-8.4 (\pm 2.2)$ & $5419.4 (\pm 64.8)$ & $729.9 (\pm 9.1)$ \\
     \bottomrule
 \end{tabular}}
 \vspace{-0.5cm}
\end{table}

\begin{table*}[ht]
\caption{Comparison between the average return of the trained policy vs that of the expert policy using $\{1, 4, 16\}$ trajectories for 1M environment steps. The table reports the mean and standard deviation of returns for \fix{5} random seeds which were evaluated using 20 episodes. The bold number indicates the best score among each algorithm.}
\label{tab:experiments_trajectories}
\centering
{\small \resizebox{\textwidth}{!}{\begin{tabular}{c| c r r r r r r r r}
    \toprule
    \# Exp Traj & Method & HalfCheetah & Ant & Walker-2D & Hopper & Humanoid & \fix{Panda-Lift} & \fix{Panda-Door} \\
    \midrule
    \multirow{7}{2em}{1} & BC & \fix{$-234.5 (\pm 171.3)$} & \fix{$-1133.0 (\pm 532.2)$} & \fix{$441.2 (\pm 321.1)$}  & \fix{$2637.2 (\pm 666.6)$}  & \fix{$2191.4 (\pm 2134.2)$} & \fix{$20.5 (\pm 16.1)$} & \fix{$206.7 (\pm 114.2)$}  \\
    & DAC & \fix{$7374.2 (\pm 57.2)$} & \fix{$4742.1 (\pm 1668.5)$} & \fix{$\mathbf{3789.2 (\pm 36.9)}$} & \fix{$3498.5 (\pm 28.9)$} & \fix{$5225.0 (\pm 275.9)$} & \fix{$\mathbf{76.8 (\pm 38.3)}$} & \fix{$260.2 (\pm 172.2)$} \\
    & OPOLO & \fix{$7357.1 (\pm 132.6)$} & \fix{$\mathbf{5604.7 (\pm 550.1)}$} & \fix{$3732.9 (\pm 162.3)$} & \fix{$3372.3 (\pm 181.8)$} & \fix{$186.3 (\pm 17.1)$} & \fix{$6.9 (\pm 2.7)$} & \fix{$\mathbf{371.0 (\pm 105.6)}$} \\
    & AIRL & \fix{$1990.8 (\pm 1077.1)$} & \fix{$-44.0 (\pm 67.2)$} & \fix{$566.2 (\pm 119.6)$} & \fix{$22.4 (\pm 13.7)$} & \fix{$186.8 (\pm 115.2)$} & \fix{$0.2 (\pm 0.1)$} & \fix{$9.9 (\pm 7.8)$} \\
    & \textit{f}-IRL & \fix{$\mathbf{7579.1 (\pm 994.4)}$} & \fix{$3230.3 (\pm 176.4)$} & \fix{$935.6 (\pm 606.6)$} & \fix{$2361.4 (\pm 1172.2)$} & \fix{$4816.8 (\pm 874.7)$} & \fix{$44.3 (\pm 26.0)$} & \fix{$0.9 (\pm 0.0)$} \\
    & OPIRL & \fix{$7416.7 (\pm 52.8)$} & \fix{$4340.5 (\pm 994.8)$} & \fix{$\mathbf{3776.4 (\pm 37.9)}$} & \fix{$\mathbf{3542.0 (\pm 27.1)}$} & \fix{$\mathbf{5331.7 (\pm 19.4)}$} & \fix{$50.5 (\pm 29.2)$} & \fix{$\mathbf{310.8 (\pm 167.6)}$} \\
    \midrule
    \multirow{6}{2em}{4} & BC & \fix{$3276.4 (\pm 1395.9)$} & \fix{$-1137.8 (\pm 714.1)$} & \fix{$2219.1 (\pm 465.3)$} & \fix{$2553.9 (\pm 1070.8)$} & \fix{$2212.8 (\pm 2068.9)$} & \fix{$173.7 (\pm 46.3)$} & \fix{$327.0 (\pm 79.3)$} \\
    & DAC & \fix{$7316.7 (\pm 73.8)$} & \fix{$5274.1 (\pm 237.9)$} & \fix{$3921.5 (\pm 92.1)$} & \fix{$2785.1 (\pm 1390.6)$} & \fix{$\mathbf{5352.8 (\pm 19.5)}$} & \fix{$115.1 (\pm 26.5)$} & \fix{$\mathbf{485.0 (\pm 0.4)}$} \\
    & OPOLO & \fix{$\mathbf{7325.6 (\pm 90.5)}$} & \fix{$\mathbf{5568.9 (\pm 298.1)}$} & \fix{$\mathbf{3787.3 (\pm 165.7)}$} & \fix{$3211.8 (\pm 245.7)$} & \fix{$249.3 (\pm 100.2)$} & \fix{$29.5 (\pm 22.1)$} & \fix{$\mathbf{485.1 (\pm 0.7)}$} \\
    & AIRL & \fix{$1751.4 (\pm 1157.6)$} & \fix{$56.9 (\pm 41.0)$} & \fix{$528.1 (\pm 97.2)$} & \fix{$13.0 (\pm 12.8)$} & \fix{$151.1 (\pm 63.7)$} & \fix{$1.0 (\pm 0.6)$} & \fix{$6.0 (\pm 5.2)$} \\
    & \textit{f}-IRL & \fix{$7049.0 (\pm 733.2)$} & \fix{$3124.4 (\pm 241.2)$} & \fix{$1947.5 (\pm 560.5)$} & \fix{$3061.3 (\pm 317.7)$} & \fix{$2905.3 (\pm 2008.2)$} & \fix{$17.2 (\pm 11.4)$} & \fix{$1.0 (\pm 0.1)$} \\
    & OPIRL & \fix{$7187.9 (\pm 200.5)$} & \fix{$2753.1 (\pm 2618.0)$} & \fix{$\mathbf{3809.6 (\pm 100.5)}$} & \fix{$\mathbf{3559.3 (\pm 26.4)}$} & \fix{$\mathbf{5264.9 (\pm 99.7)}$} & \fix{$\mathbf{254.7 (\pm 107.6)}$} & \fix{$\mathbf{477.6 (\pm 10.7)}$} \\
    \midrule
    % N_trajectories = 16
    \multirow{6}{2em}{16} & BC & \fix{$7187.8 (\pm 132.1)$} & \fix{$4338.5 (\pm 475.0)$} & \fix{$3797.6 (\pm 37.2)$} & \fix{$3550.5 (\pm 8.9)$} & \fix{$4927.0 (\pm 339.5)$} & \fix{$350.8 (\pm 36.1)$} & \fix{$\mathbf{481.4 (\pm 6.6)}$} \\
    & DAC & \fix{$7232.0 (\pm 135.1)$} & \fix{$5405.4 (\pm 235.8)$} & \fix{$\mathbf{3845.7 (\pm 81.0)}$} & \fix{$3523.4 (\pm 22.6)$} & \fix{$5281.1 (\pm 78.4)$} & \fix{$242.0 (\pm 164.2)$} & \fix{$\mathbf{483.6 (\pm 3.5)}$} \\
    & OPOLO & \fix{$\mathbf{7353.5 (\pm 111.6)}$} & \fix{$\mathbf{5714.3 (\pm 340.3)}$} & \fix{$3651.6 (\pm 151.3)$} & \fix{$3182.2 (\pm 449.4)$} & \fix{$518.9 (\pm 235.7)$} & \fix{$124.7 (\pm 71.7)$} & \fix{$\mathbf{485.6 (\pm 0.3)}$} \\
    & AIRL & \fix{$1356.8 (\pm 1152.2)$} & \fix{$21.7 (\pm 19.8)$} & \fix{$668.9 (\pm 140.8)$} & \fix{$31.0 (\pm 39.7)$} & \fix{$159.4 (\pm 68.9)$} & \fix{$1.4 (\pm 1.5)$} & \fix{$11.7 (\pm 8.4)$} \\
    & \textit{f}-IRL & \fix{$6866.3 (\pm 646.3)$} & \fix{$3006.2 (\pm 562.4)$} & \fix{$1405.2 (\pm 688.2)$} & \fix{$3126.1 (\pm 244.6)$} & \fix{$4372.7 (\pm 1607.7)$} & \fix{$15.2 (\pm 8.6)$} & \fix{$1.0 (\pm 0.0)$} \\
    & OPIRL & \fix{$7263.5 (\pm 97.1)$} & \fix{$5063.1 (\pm 462.4)$} & \fix{$3766.0 (\pm 54.7)$} & \fix{$\mathbf{3672.6 (\pm 130.7)}$} & \fix{$\mathbf{5359.2 (\pm 13.6)}$} & \fix{$\mathbf{429.3 (\pm 39.1)}$} & \fix{$\mathbf{479.1 (\pm 9.8)}$} \\
    \midrule
    \midrule
    Expert & SAC & $6991.7 (\pm 124.3)$ & $4875.1 (\pm 1036.8)$ & $3665.1 (\pm 527.2)$ & $3298.7 (\pm 433.2)$ & $5277.2 (\pm 490.7)$ & \fix{$468.4 (\pm 12.8)$} & \fix{$485.4 (\pm 0.3)$} \\
    \bottomrule
\end{tabular}}}
\vspace{-2mm}
\end{table*}

The first task is evaluated on a 2D maze environment, where we control a point mass to navigate from start (yellow ball) to goal (green ball) while avoiding a barrier.
The position of the barrier changes from left to right on test time (see Fig.\ref{fig:3_a} and Fig.~\ref{fig:3_b}) .
The results in Table \ref{tab:transfer_learning} show that only OPIRL successfully solves the task in the target environments.
Other IRL algorithms could not achieve a policy with near-optimal performance when the environment structure changed compared to our method.
This was caused by multiple factors such as insufficient number of interactions and not reconstructing optimal reward functions during training on the source environment.
On the other hand, IL methods perform poorly in reward robustness tasks.

For the second task, we experiment on a quadrupedal ant agent, which is a modified version of OpenAI's Gym Ant-v2. 
During reward training, the agent is trained on a regular quadruped-ant environment (see Fig.~\ref{fig:3_c}), then we modify the agent in two ways.
The BigAnt (see Fig.~\ref{fig:3_d}) is an environment where the length of all legs are doubled.
The Amputated Ant (see Fig.~\ref{fig:3_e}), on the other hand, has two shortened front legs that can significantly change the gait.
As seen in Table \ref{tab:transfer_learning}, OPIRL successfully recovers a near-optimal performance compared to other IRL methods.
IL methods, which require direct policy generalization, fail to move forward in either task, even though they achieve near-optimal performance on the source environment. 
This is due to over-fitting to the trained environment and not learning the underlying goal of the new task.

% %%%%%%%%%%%%%%%%%%%%%%%%%%%%%%%%%%%%%%%%%%%%%%%%%%%%%%%%%%%%%%%%%%%%%%%%%%%%%%%%
\section{Ablation Experiments} \label{sec:ablation}

In this section, we perform a series of ablation experiments to understand what components contribute to the performance gain.

\subsection{Effects of Behavior Cloning Loss and Q-Filter} \label{subsec:ablation_reg}

We investigate the effect of additional regularization to the policy.
We've performed further ablation studies on 2 new variants: "OPIRL no reg" indicates OPIRL without either BC Loss or Q-Filter, and "OPIRL no Q-Filter" indicates OPIRL without Q-Filter.
We observe the effect of regularization, as seen in Fig.~\ref{fig:sample_efficiency}.
Without Q-Filter, the agent fails to choose optimal action, leading to slower convergence.
Applying both BC Loss and Q-Filter will have the best effect where other variants require more training time.

\subsection{Number of Trajectories} \label{subsec:ablation_n_traj}

We run experiments on multiple trajectories to observe the difference in performance.
For locomotion tasks, when we provide additional expert demonstration, all three off-policy based methods, DAC, OPOLO, and OPIRL, show consecutive high performance.
BC also shows an increase in performance when provided with ample expert trajectories, a behavior seen in prior work \cite{ni_f-irl_2020}.
AIRL, however, cannot achieve high performance even with 16 expert trajectories due to lack of environment interactions (1M steps).
{\it f}-IRL, although showing decent performance on all tasks, could not reach expert-level on complicated tasks such as the robotic tasks.
For robotic tasks, due to lack of demonstration in a single trajectory, all methods show difficulty in achieving optimality.

% %%%%%%%%%%%%%%%%%%%%%%%%%%%%%%%%%%%%%%%%%%%%%%%%%%%%%%%%%%%%%%%%%%%%%%%%%%%%%%%%
\section{Conclusion} \label{sec:conclusion}
Imitation learning and inverse reinforcement learning have eliminated hand-crafted reward design, which requires tremendous effort or is even infeasible in complex systems. However, we have seen that prior works of the former cannot generalize to different environments, and the latter requires a massive number of interactions with environments.
In this paper, we propose OPIRL that has the two desired properties: high sample efficiency and generalization capability to unseen environments.
To achieve this, OPIRL combines three main components: 1) adopts off-policy data distribution, 2) leverages mode-covering behavior for faster training, and 3) learns a stationary reward function.
Our experiments demonstrated that OPIRL can achieve comparable or better sample efficiency among state-of-the-art IL and IRL methods with the limited data regime. In addition, it achieves the highest performance on novel environments.

In the future, we plan to expand the usability of our algorithm to more complicated settings, such as high dimensional inputs (\textit{e.g.,} images, videos). We also aim to investigate how we can make use of the proposed method for controlling real systems.

\printbibliography %addeda

\appendices

\section{Proof} 
\subsection{Eq.~(3): Derivation of KL-Divergence Upper-bound} \label{sec:kl_divergence_upper_bound_proof}

Inspired by the works of \cite{nachum_algaedice_2019, zhu_off-policy_2021} we set an upper-bound on KL-divergence using {\it f}-divergence. 
For two arbitrary distributions $P$ and $Q$, using the Jensen's inequality, the upper bound of KL-divergence can be shown as $D_{KL}[P \| Q] \leq D_{f} [P \| Q]$:
\begin{eqnarray}
D_{KL}[P \| Q] &=& \int p(x) \log \frac{p(x)}{q(x)} dx \nonumber \\
&\leq& \log \int  \frac{p^2(x)}{q(x)} dx \nonumber \\
&\leq& \frac{1}{2} \int  \frac{p^2(x)}{q(x)} dx = D_f [P \| Q]
\end{eqnarray}
where $f(x) = \frac{1}{p} |x|^{p}$.
From prior empirical results \cite{nachum_algaedice_2019}, in our work we use $f(x) = \frac{1}{2} x^{3/2}$ for a tighter upper bound.

\subsection{Eq.~(11): Derivation}
Below we show the steps we took to derive Eq.~(11):
\begin{eqnarray*}
    &&\mathrm{KL}\left(\rho^{\pi_{\theta}}(s, a) \| \rho^{\exp }(s, a)\right) \\
    &=& \int \distagent(s, a) \log \left(\frac{\distagent(s, a)}{\distexp(s, a)} \right) ds da\\
    &=& \int \distagent(s, a) \log \left( \frac{\distagent(s, a)}{\distrb (s, a} \frac{\distrb (s, a)}{\distexp (s, a)} \right) ds da \\
    &=& \int \distagent (s, a) \log \left( \frac{\distagent(s, a)}{\distrb (s, a} \right) ds da \\
    && \quad + \int \distagent (s, a) \log \left( \frac{\distrb (s, a)}{\distexp (s, a)} \right) ds da \\
    &=& \mathbb{E}_{\rho^{\pi_{\theta}}}\left[\log \frac{\distrb (s, a)}{\distexp (s, a)}\right]+\mathrm{KL}\left(\distagent (s, a) \| \distrb (s, a)\right)
    % \quad=
\end{eqnarray*}

\subsection{Eq.~(13): Derivation}

The variational form of {\it f}-divergence can be expressed as:
\begin{equation*} \label{eq:f_divergence_variational}
\begin{split}
    - &\mathbb{D}_f [\distagent (s, a) \| \distrb (s, a) ] \\
     &= \inf_{x: \state \times \action \rightarrow R} \mathbb{E}_{\distagent} [-x (s, a)] + \mathbb{E}_{\distrb} [f_\ast (x (s, a))].           
    \end{split}
\end{equation*}

Using this notion, we apply this back into Eq.~(12):

\begin{eqnarray*}
    && \mathbb{E}_{\distagent}\left[\log \frac{\distrb (s, a)}{\distexp (s, a)}\right]+\mathbb{D}_{f}\left[\distagent (s, a) \| \distrb (s, a)\right] \\
    &=& \mathbb{E}_{\distagent}\left[\log \frac{\distrb (s, a)}{\distexp (s, a)}\right]
    - \mathbb{E}_{\distagent} [x(s, a)] + \mathbb{E}_{\distrb} [f_\ast (x(s, a))] \\
    &=& \mathbb{E}_{\distagent}\left[\log \frac{\distrb (s, a)}{\distexp (s, a)} - x (s, a) \right] + \mathbb{E}_{\distrb} [f_\ast (x(s, a))]
\end{eqnarray*}

\subsection{Derivation of our Eq.(14)}

We show the equation transformation from on-policy training to off-policy training in Eq.~(12) in the main paper.
Recall that the Bellman equation for state-action Q-function is $\mathcal{B}^{\pi}Q(s,a) = \mathbb{E}_{s'\sim P(s,a),a'\sim \pi(s^\prime)}\left[r(s,a) + \gamma Q(s',a')\right]$, and $r(s, a) = \log \frac{\distexp (s, a)}{\distrb (s, a)}$.
The RHS of the equation
\begin{equation}
\begin{split}
J(\pi, Q) =& \mathbb{E}_{\distagent}\left[\log \frac{\distexp \left(s, a\right)}{\distrb \left(s, a\right)}-\left(\mathcal{B}^{\pi} Q-Q\right)(s, a)\right] \\
&+\mathbb{E}_{\distrb}\left[f_{*}\left(\left(\mathcal{B}^{\pi} Q-Q\right)(s, a)\right)\right]
\end{split}
\end{equation}
can be expressed via initial state and actions in the following manner:
\begin{figure*}[t]
    \centering
    \includegraphics[width=\linewidth]{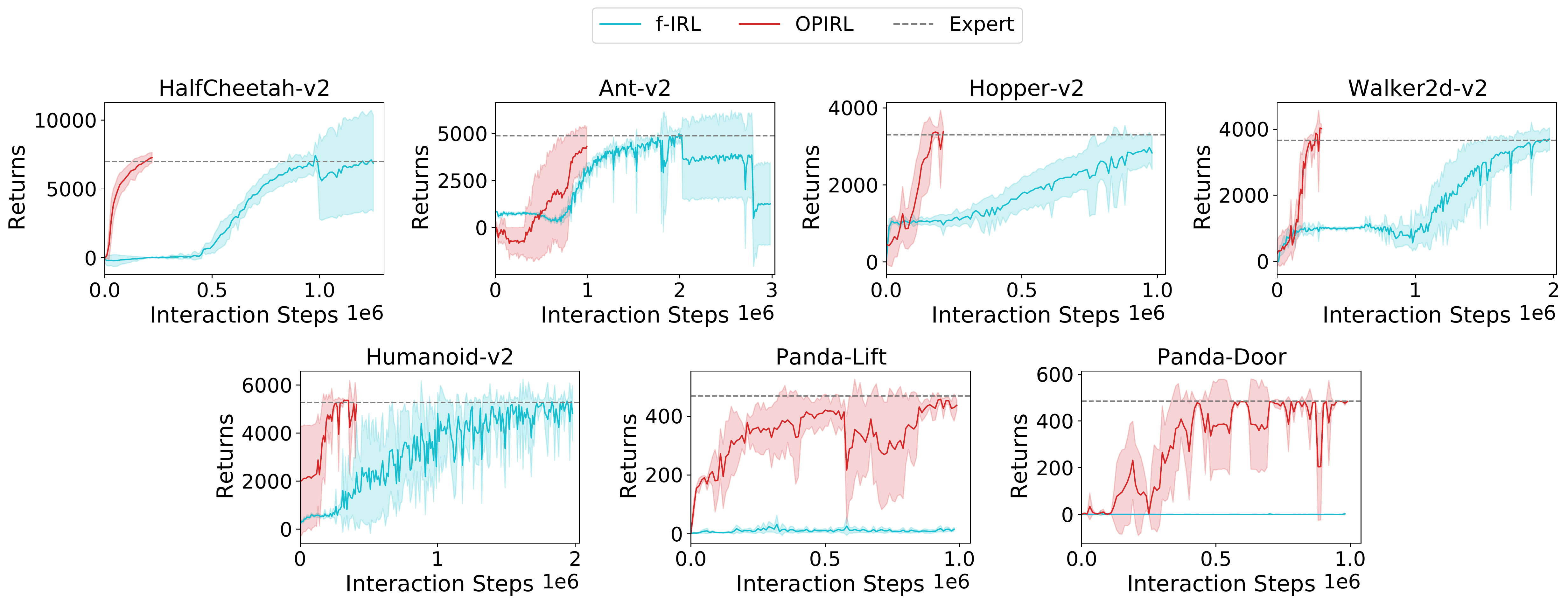}
    \caption{Comparison between our method and {\it f}-IRL. We show the training curve until it reaches the expert level for couple of steps. OPIRL reaches expert-level performance in significantly less steps than {\it f}-IRL.}
    \label{fig:diff_firl}
\end{figure*}

\begin{eqnarray}
 && \mathbb{E}_{\distagent}\left[\log \frac{\distexp \left(s, a\right)}{\distrb \left(s, a\right)}-\left(\mathcal{B}^{\pi} Q-Q\right)(s, a)\right] \nonumber  \\
 &\Leftrightarrow& \mathbb{E}_{\distagent} \left[ r(s, a) - Q(s, a) - \mathbb{E}_{s' \sim P(s, a)} [ \mathcal{B}^{\pi} Q(s, a)\right] \nonumber \\
 &\Leftrightarrow& \mathbb{E}_{\distagent} [ r(s, a) + Q(s, a) \nonumber \\
 && \quad \quad \quad - \mathbb{E}_{s' \sim P(s, a), a' \sim \pi(s)} [r(s, a) + \gamma Q(s' , a')]] \nonumber \\
&\Leftrightarrow& \mathbb{E}_{\distagent} \left[ Q(s, a) - \gamma \mathbb{E}_{s' \sim P(s, a), a' \sim \pi(s^\prime)} [ Q(s' , a')]\right]  \nonumber \\
&\Leftrightarrow& (1 - \gamma) \sum_{t=0} ^{\infty} \gamma ^t \mathbb{E}_{s \sim \distagent_t, a \sim \pi(s)} \left[ Q(s, a)\right] \nonumber \\
&& - (1 - \gamma) \sum_{t=0} ^{\infty} \gamma ^{t+1} \mathbb{E}_{s' \sim P(s, a), a'  \sim \pi(s')} \left[ Q(s', a')\right] \nonumber \\
&\Leftrightarrow& (1 - \gamma) \sum_{t=0} ^{\infty} \gamma^t \mathbb{E}_{s \sim \distagent_t, a \sim \pi(s)} \left[Q(s, a) \right] \nonumber \\
&& - (1 - \gamma) \sum_{t=0} ^{\infty} \gamma ^{t+1} \mathbb{E}_{s' \sim \distagent_{t+1}, a^\prime \sim \pi(s')} \left[ Q(s', a') \right] \nonumber \\
&\Leftrightarrow& (1 - \gamma) \mathbb{E}_{s_0 \sim \distagent_0, a_0 \sim \pi(s_0)} \left[ Q(s_0, a_0) \right].
\end{eqnarray}
Therefore, combining it with LHS, we get the objective function defined in Eq.~(12):
\begin{equation}
\begin{split}
\max_\pi & \min_{x: \mathcal{S} \times \mathcal{A} \rightarrow R} J(\pi, x) \\
&= \max_\pi \min_{Q: \state \times \action \rightarrow R} J(\pi, Q) \\
&= (1 - \gamma) \mathbb{E}_{s_0 , a_0} [Q(s_0, a_0)] + \mathbb{E}_{\distrb}[f_\ast ((\mathcal{B}^\pi Q - Q)(s, a))].
\end{split}
\end{equation}

\section{Experiment Details} \label{sec:experiment_details}

\begin{figure}[h]
    \centering
     \begin{minipage}[b]{0.32\linewidth}
        \captionsetup{justification=centering}
        \centering
        \subfloat[HalfCheetah-v2]{\label{fig:7_a}\includegraphics[width=0.95\linewidth]{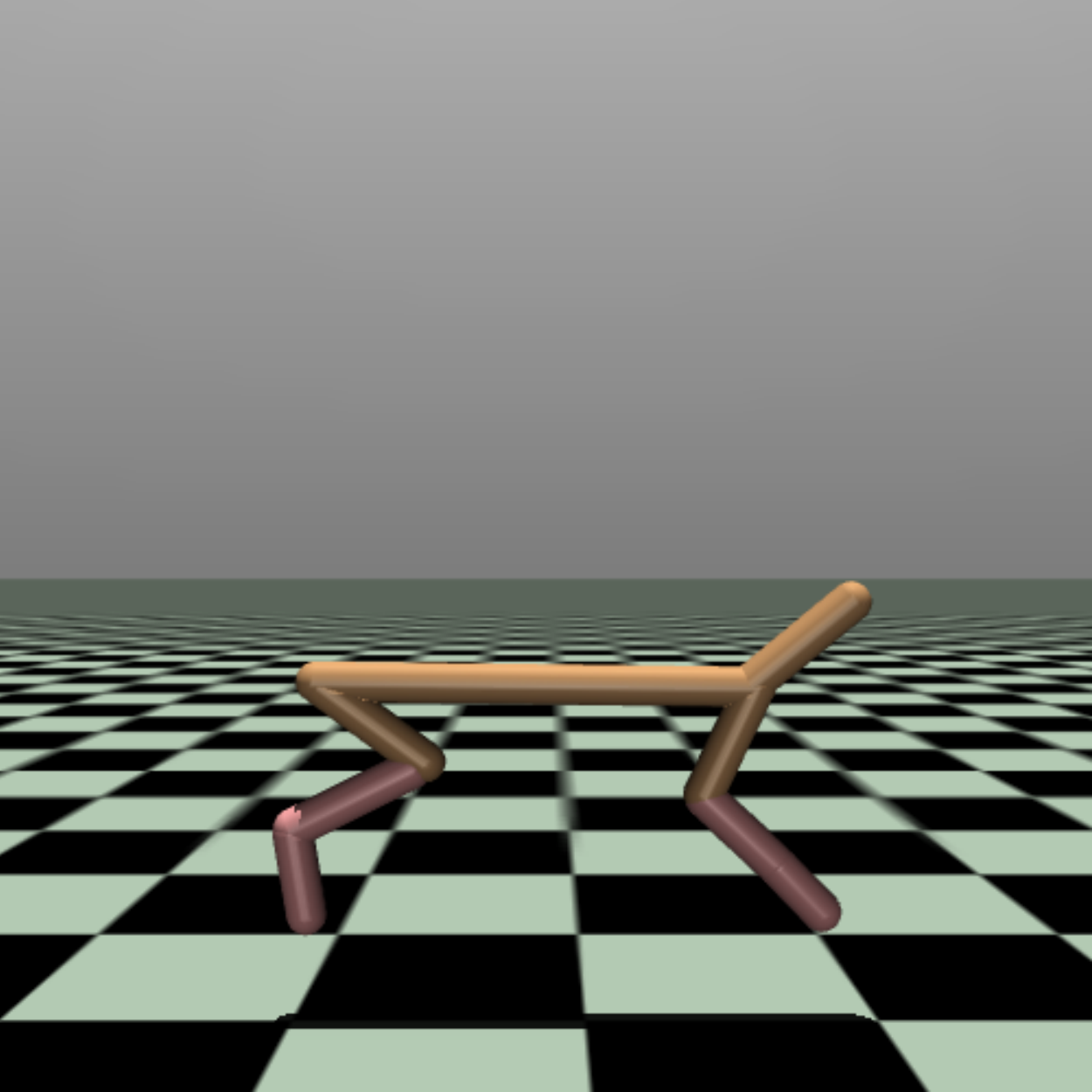}}
     \end{minipage}
     \begin{minipage}[b]{0.32\linewidth}
        \captionsetup{justification=centering}
         \centering
        \subfloat[Ant-v2]{\label{fig:7_b}\includegraphics[width=0.95\linewidth]{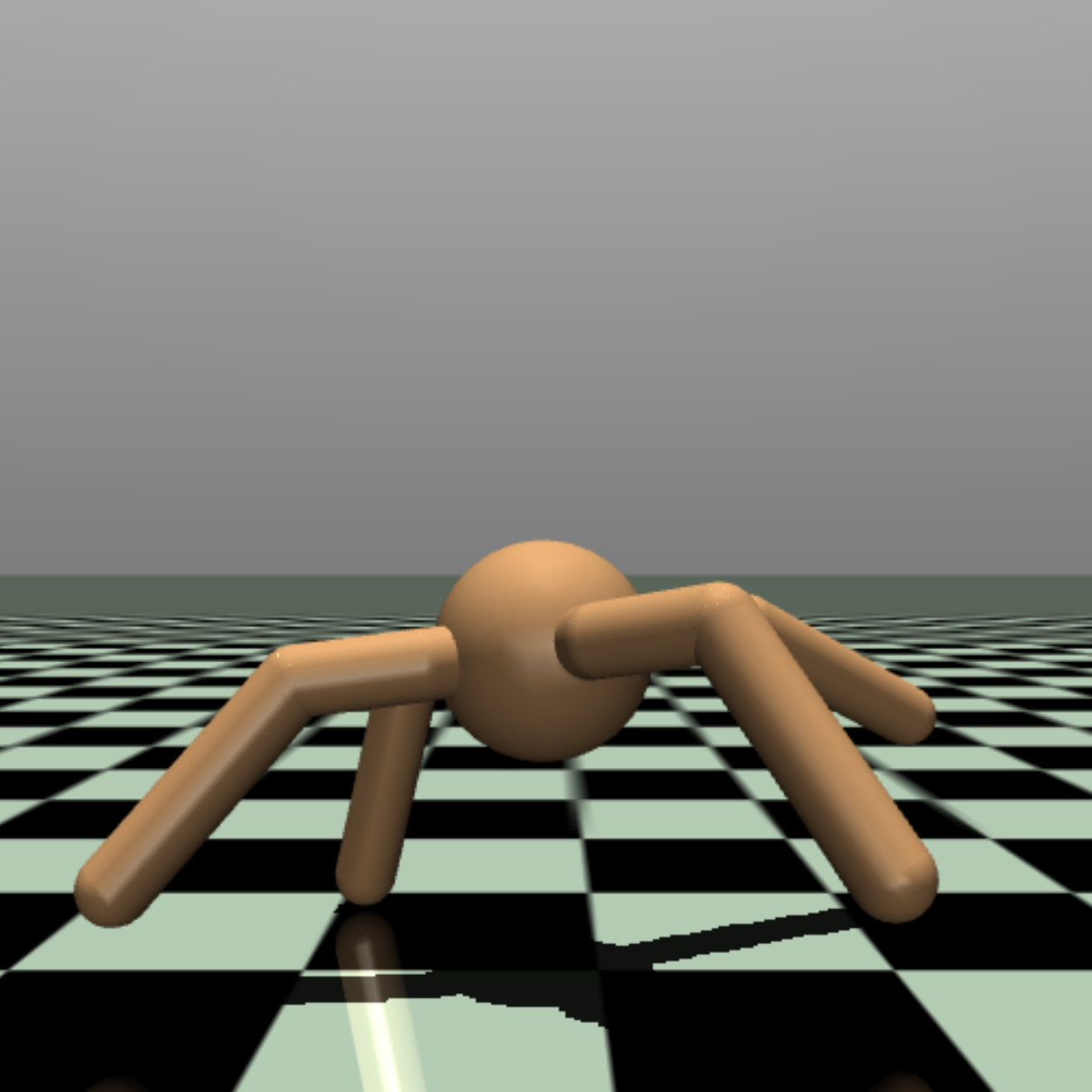}}
     \end{minipage}
    \begin{minipage}[b]{0.32\linewidth}
        \captionsetup{justification=centering}
        \centering
        \subfloat[Walker2d-v2]{\label{fig:7_c}\includegraphics[width=0.95\linewidth]{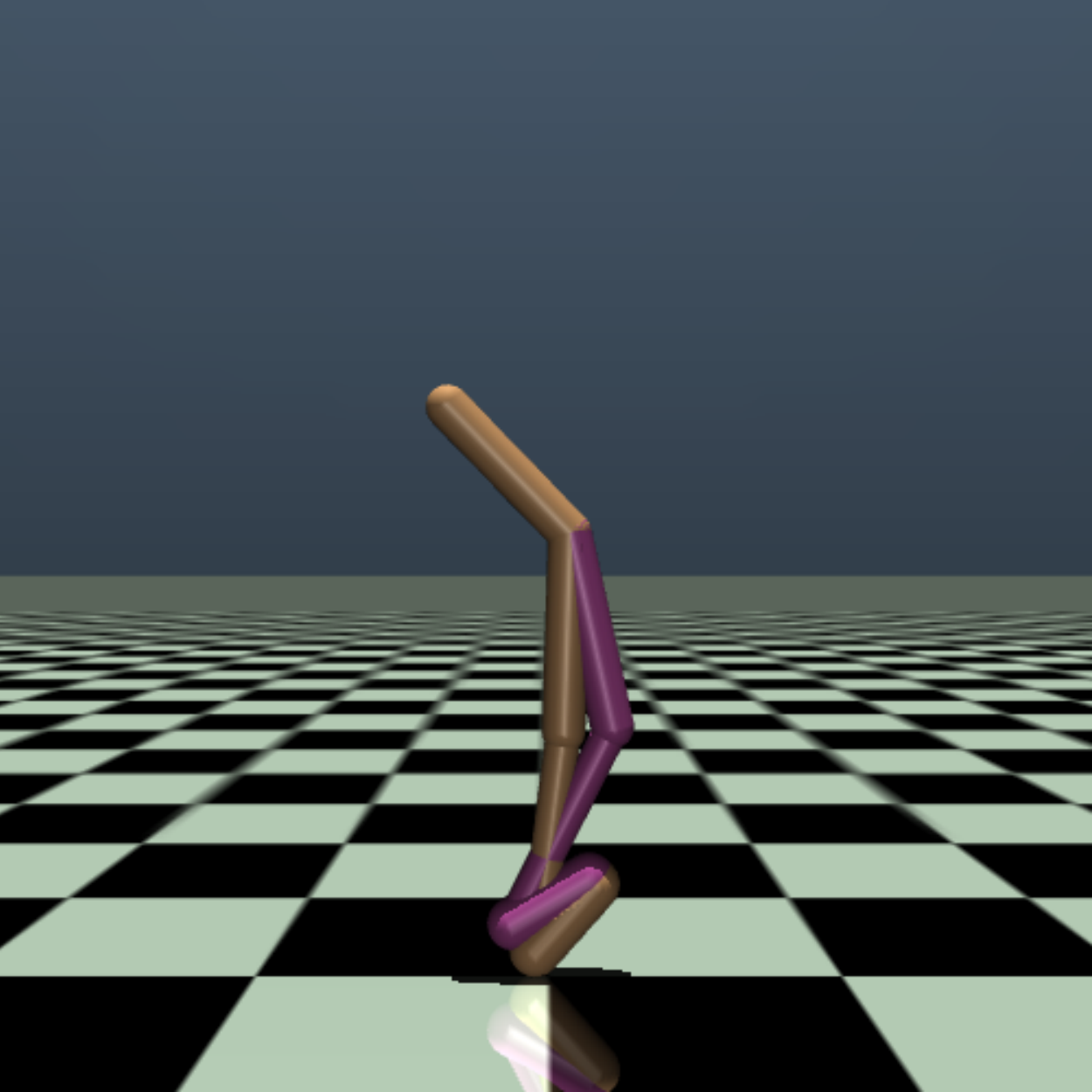}}
    \end{minipage}  
    \begin{minipage}[b]{0.32\linewidth}
        \captionsetup{justification=centering}
        \centering
        \subfloat[Hopper-v2]{\label{fig:7_d}\includegraphics[width=0.95\linewidth]{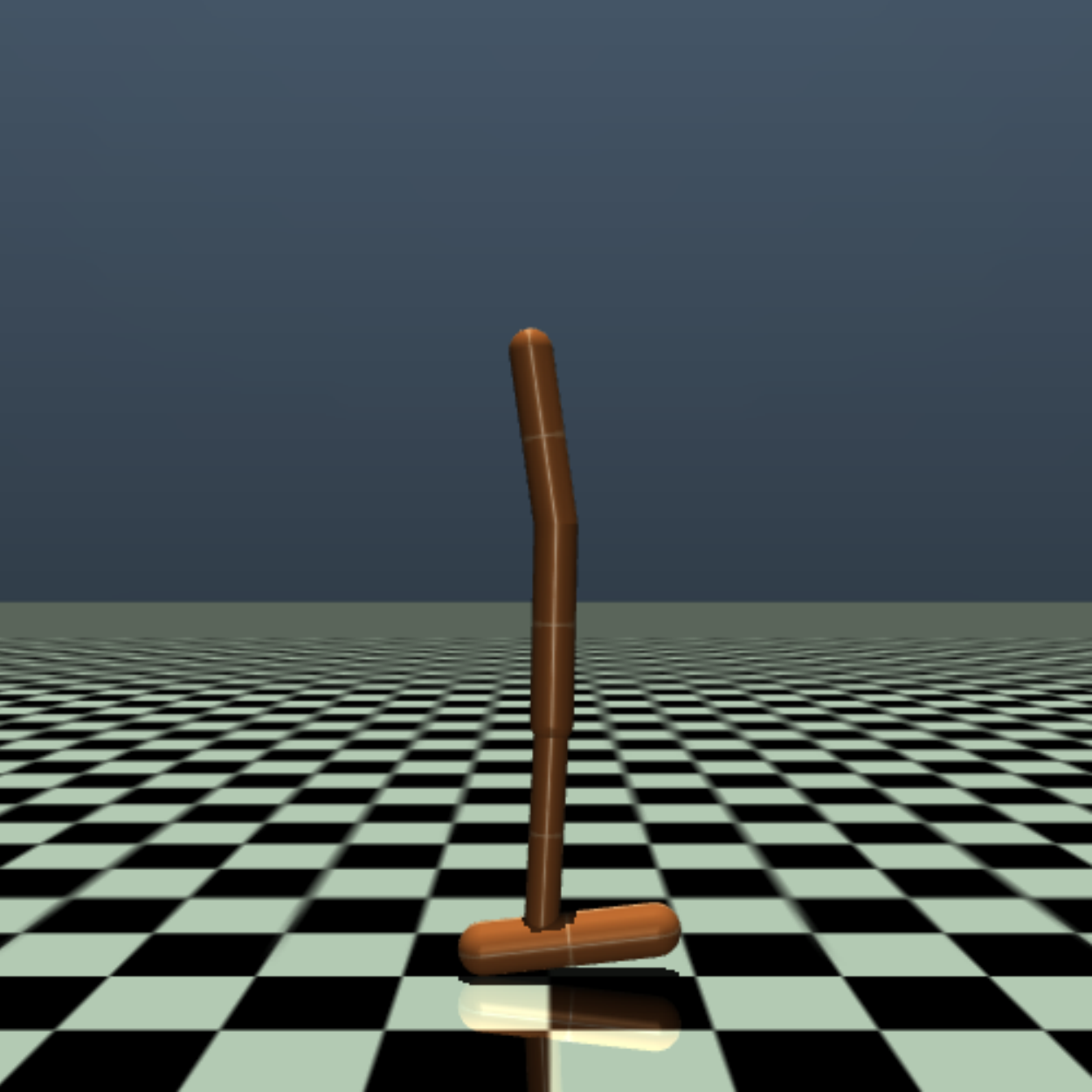}}
    \end{minipage}   
    \begin{minipage}[b]{0.32\linewidth}
        \captionsetup{justification=centering}
        \centering
        \subfloat[Humanoid-v2]{\label{fig:7_e}\includegraphics[width=0.95\linewidth]{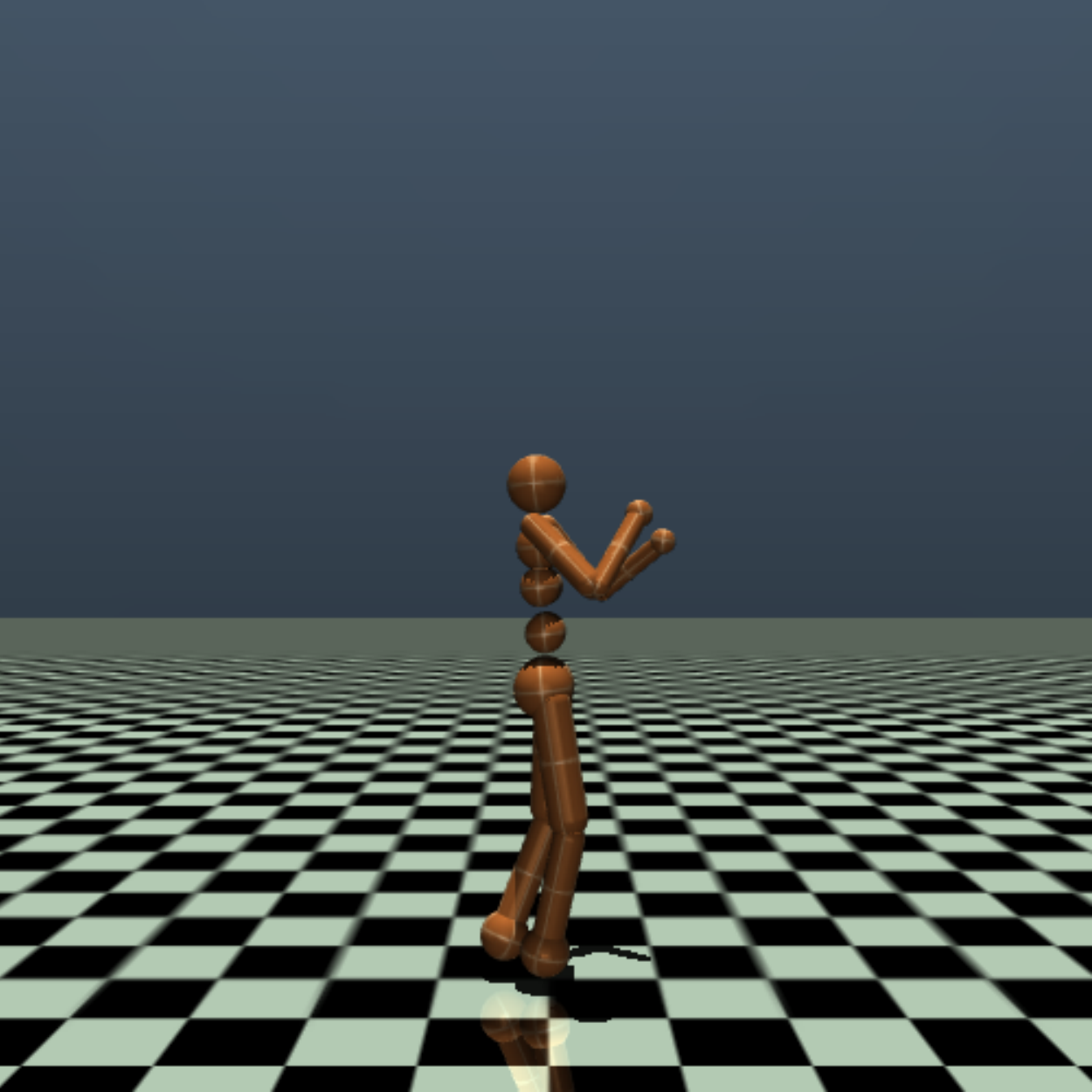}}
    \end{minipage}    
    \begin{minipage}[b]{0.32\linewidth}
        \captionsetup{justification=centering}
        \centering
        \subfloat[Panda-Lift]{\label{fig:8_e}\includegraphics[width=0.95\linewidth]{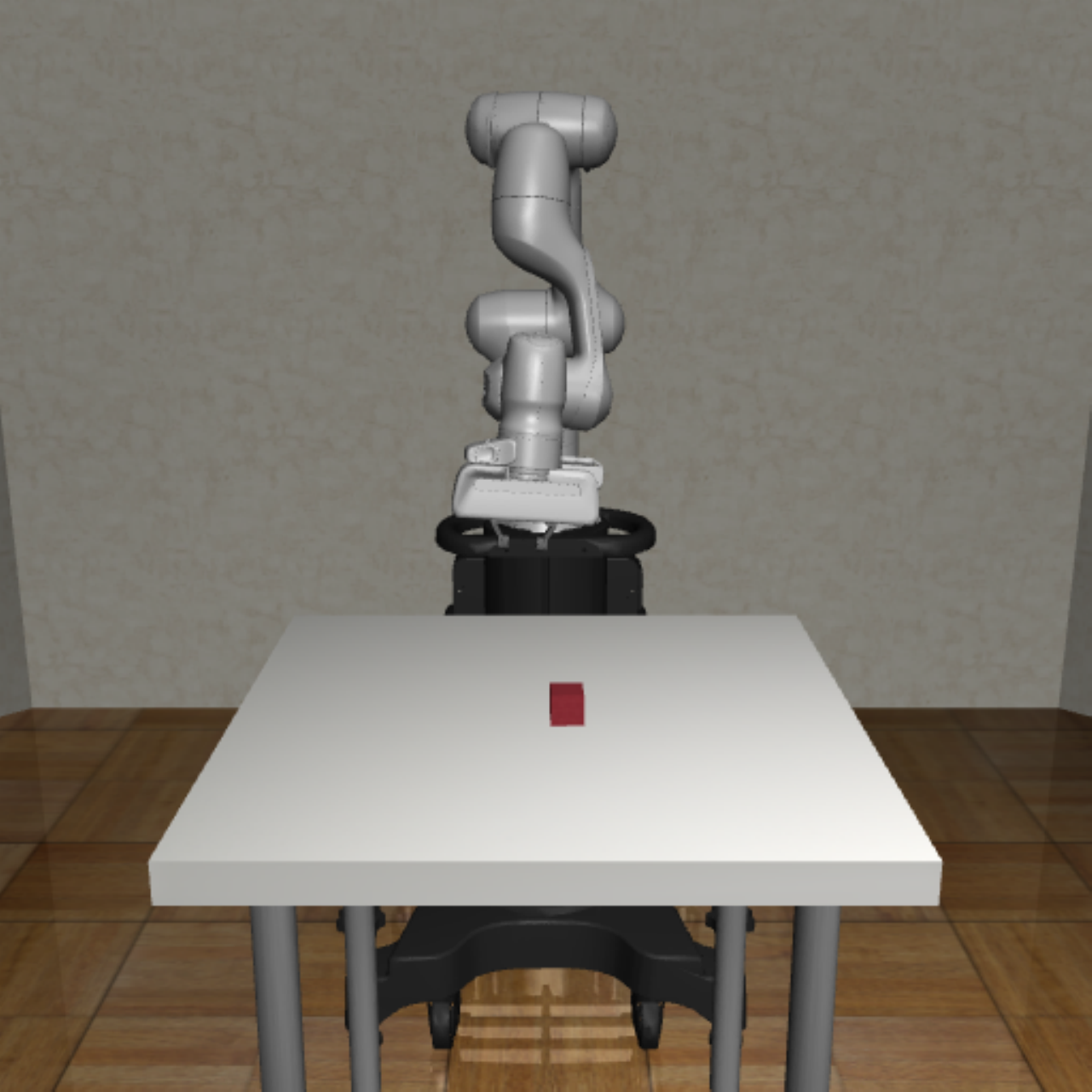}}
    \end{minipage}    
    \begin{minipage}[b]{0.32\linewidth}
        \captionsetup{justification=centering}
        \centering
        \subfloat[Panda-Door]{\label{fig:9_e}\includegraphics[width=0.95\linewidth]{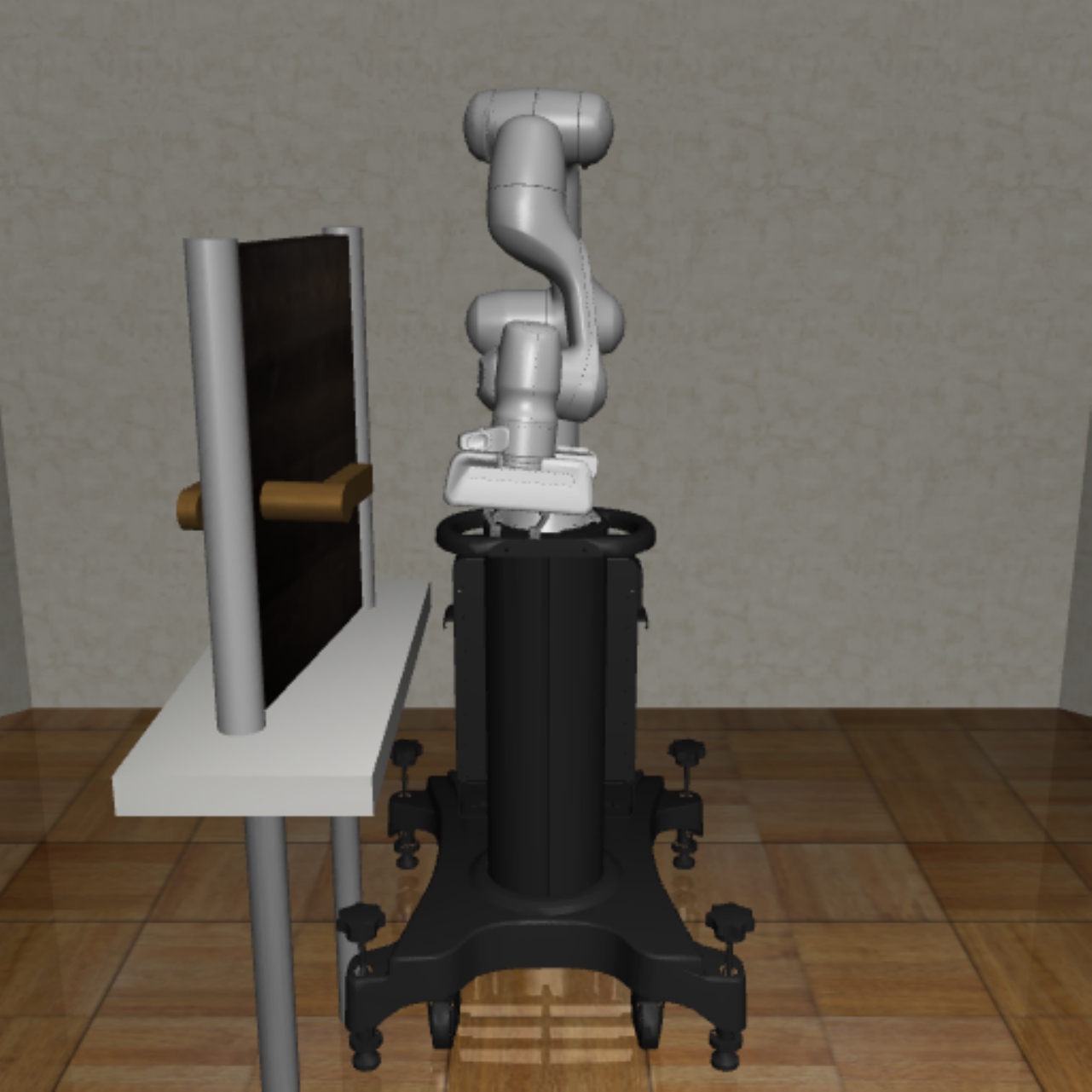}}
    \end{minipage}    
    \caption{Environments used in policy performance evaluation tasks.}
    \label{fig:mujoco_env}
\end{figure}
In this section, we provide details on our experiments. 
\subsection{Implementation Details}
All networks are structured using a 2-layer fully-connected network.
For reward functions we use 64 hidden units with ReLU activation~\cite{nair_rectified_2010}, and 256 for other networks.
All networks uses Adam optimizer~\cite{kingma_adam_2017} with $10^{-5}$ for actor and discriminator network, and $10^{-3}$ for critic network for all tasks except for HalfCheetah-v2 where the discriminator uses $3 \times 10^{-4}$.
The actor loss regularization coefficient and behavior cloning loss regularization coefficient uses $10^{-3}$ and $1 / \mathrm{batch\_size}$ respectively, where the batch size is set to $256$ for all tasks.
We use absorbing states of the environments following the works of Kostrikov et at. \cite{kostrikov_discriminator-actor-critic_2018} and normalize states for stable training.
Therefore, the replay buffer size is set to $2 \times \mathrm{total timesteps}$ where the total timesteps is set to $10^6$ for all policy performance tasks.
For transfer learning, we train the reward functions for $5 \times 10^5$ to avoid over-fitting to the current policy.
Furthermore, we used $16$ trajectories for PointMaze task to increase the variety of goal position.

\fix{We use gradient penalty \cite{gulrajani_improved_2017} to enhance stable training for discriminators in reward training.
Other options for discriminator regularization techniques include spectral normalization \cite{miyato_spectral_2018}, Mixup \cite{zhang_mixup_2018, chen_batch_2021}, and PUGAIL \cite{xu_learning_2019}, however we chose GP as it has been empirically shown to achieve decent performance across multiple tasks \cite{orsini_what_2021, blonde_lipschitzness_2021}.}
    
In our policy update, following the technique used in AlgaeDICE \cite{nachum_algaedice_2019}, we mix the Q-value function $Q_\phi(s', a')$ in Eq.~(16) with target Q-value $\bar{Q}_\phi (s', a')$ as follows:
\begin{equation}
    \lambda_3 Q_\phi (s', a') + (1 - \lambda_3) \bar{Q}_\phi (s', a').
\end{equation}
where $\lambda_3$ is set to $0.05$.
All expert demonstrations were collected via training an agent using Soft-Actor-Critic \cite{haarnoja_soft_2018} using the library TF2RL\footnote{\url{https://github.com/keiohta/tf2rl/}}~\cite{ota_tf2rl_2020}.

We emphasize that our method requires minimal hyperparamter tuning.
In contrast, prior arts such as AIRL, require excessive tuning, and minor difference can cause the training to fail.

For OPOLO \cite{zhu_off-policy_2021} and {\it f}-IRL \cite{ni_f-irl_2020}, we use its original implementations and hyperparameters.

\subsection{Environments}
For policy performance tasks, we use 5 continual control tasks from OpenAI Gym \cite{brockman_openai_2016} simulated on a physics simulator MuJoCo \cite{todorov_mujoco_2012}: HalfCheetah-v2, Ant-v2, Walker2d-v2, Hopper-v2, and Humanoid-v2 (See Fig.~{\ref{fig:mujoco_env}}) \fix{, and 2 robotic tasks from Robosuite \cite{zhu_robosuite_2020}: Lift and Door}.

Environments for transfer learning tasks involves the point-mass environment, and quadrupedal ant environment introduced in the original AIRL paper \cite{fu_learning_2018}.
In addition, we use an extension of the quadrupedal ant environment, where the length of all legs are double, which was also done in prior works \cite{qureshi_adversarial_2019}.

\subsection{Transfer Learning Visualization Results}
\begin{figure}
    \begin{minipage}[b]{0.49\linewidth}
       \centering
       \includegraphics[width=\linewidth]{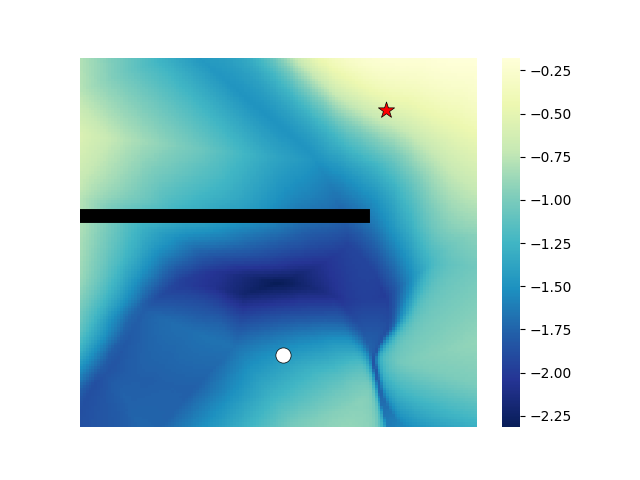}
    \end{minipage}
    \begin{minipage}[b]{0.49\linewidth}
    \centering
    \includegraphics[width=\linewidth]{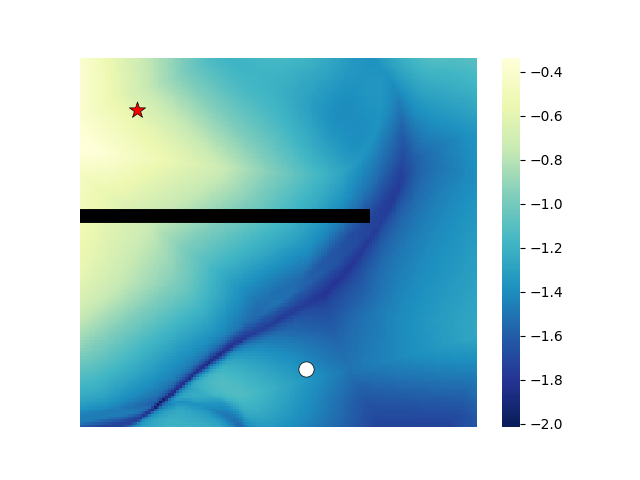}
    \end{minipage}
    \caption{Visualization of estimated reward function by using OPIRL. The white circle represents the agent, the red star represents the goal position, and the black rectangle shows the barrier.}
    \label{fig:pointmaze_reward}
\end{figure}

\begin{figure*}[t]
\centering
% \vspace*{-0.25cm}
\begin{tabular}{ccccc}
  \begin{minipage}{0.19\linewidth}
  \includegraphics[width=0.95\linewidth]{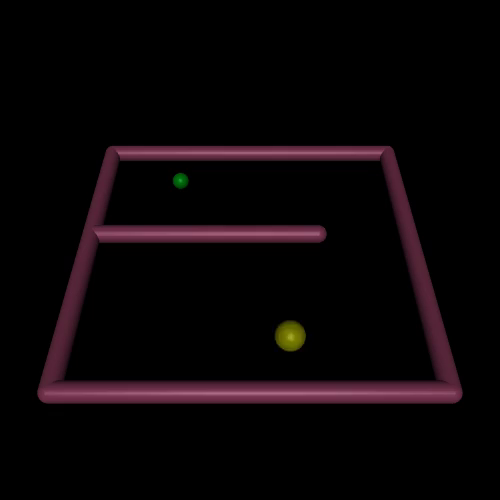}
  \vspace{0.2cm}
  \end{minipage}
  \begin{minipage}{0.19\linewidth}
  \includegraphics[width=0.95\linewidth]{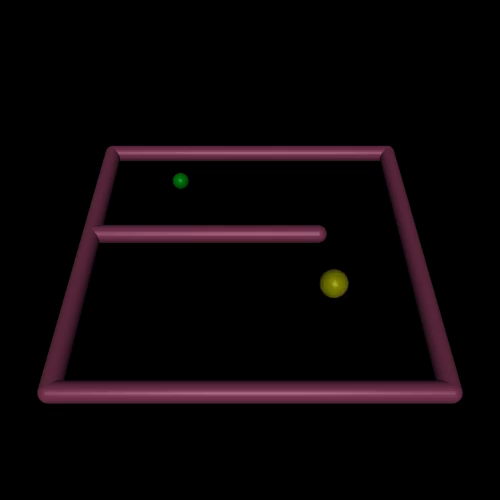}
  \vspace{0.2cm}
  \end{minipage}
  \begin{minipage}{0.19\linewidth}
  \includegraphics[width=0.95\linewidth]{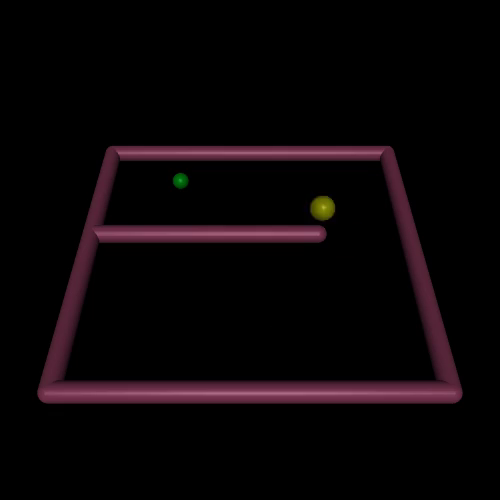}
  \vspace{0.2cm}
  \end{minipage}
  \begin{minipage}{0.19\linewidth}
  \includegraphics[width=0.95\linewidth]{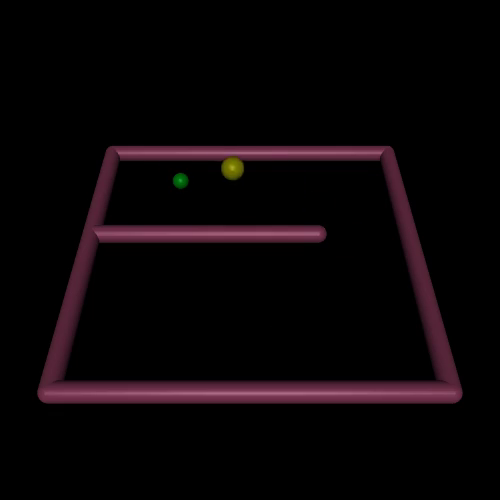}
  \vspace{0.2cm}
  \end{minipage}
  \begin{minipage}{0.19\linewidth}
  \includegraphics[width=0.95\linewidth]{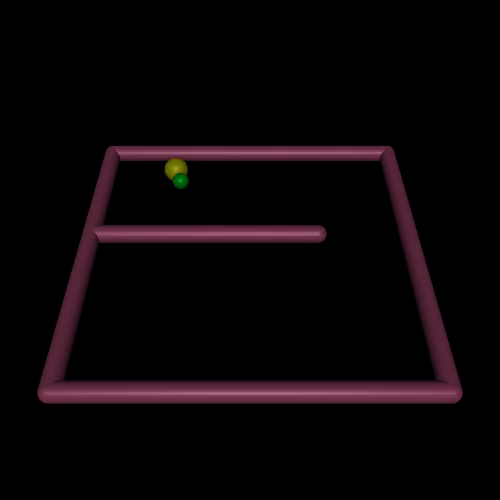}
  \vspace{0.2cm}
  \end{minipage} 
  \\
  \begin{minipage}{0.19\linewidth}
  \includegraphics[width=0.95\linewidth]{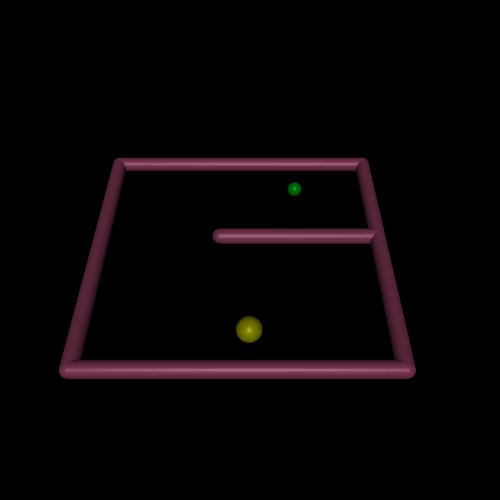}
  \vspace{0.2cm}
  \end{minipage}
  \begin{minipage}{0.19\linewidth}
  \includegraphics[width=0.95\linewidth]{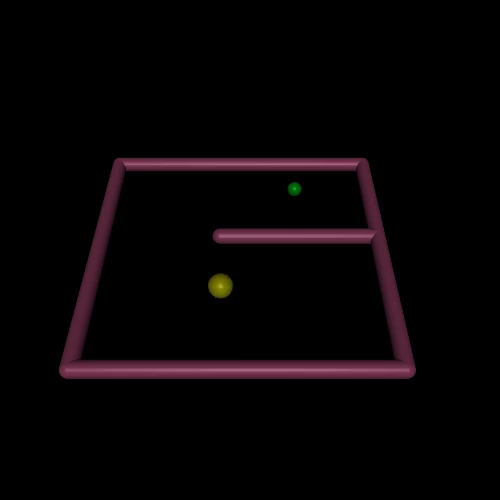}
  \vspace{0.2cm}
  \end{minipage}
  \begin{minipage}{0.19\linewidth}
  \includegraphics[width=0.95\linewidth]{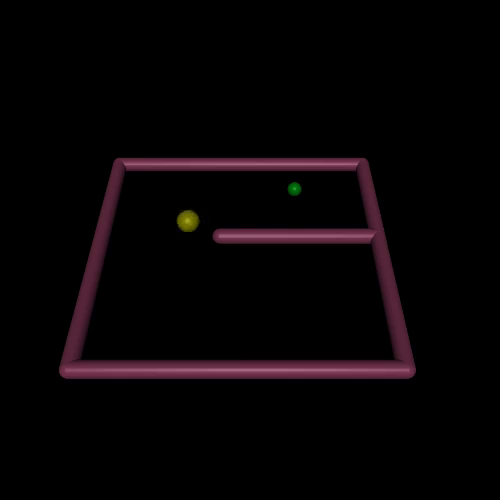}
  \vspace{0.2cm}
  \end{minipage}
  \begin{minipage}{0.19\linewidth}
  \includegraphics[width=0.95\linewidth]{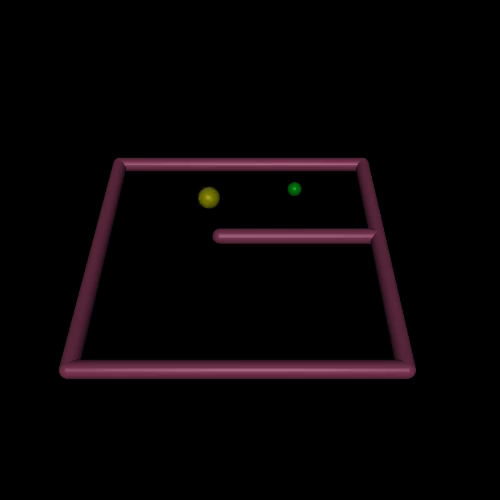}
  \vspace{0.2cm}
  \end{minipage}
  \begin{minipage}{0.19\linewidth}
  \includegraphics[width=0.95\linewidth]{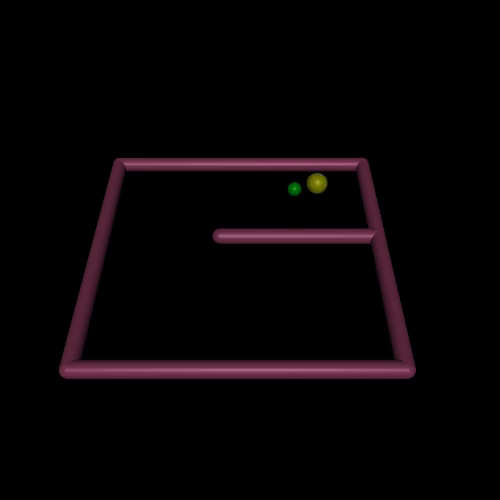}
  \vspace{0.2cm}
  \end{minipage}  
\end{tabular}
\caption{\textbf{Top row:} The original PointMaze-Left, where the ball (yellow) tries to reach the goal (green). \textbf{Bottom row:} OPIRL successfully learns a new policy on PointMaze-Right via transferring the reward learned on the source environment.}
\label{fig:transfer_point_maze}
\end{figure*}

\begin{figure*}[tb]
\centering
% \vspace*{-0.25cm}
\begin{tabular}{ccccc}
  \begin{minipage}{0.19\linewidth}
  \includegraphics[width=0.95\linewidth]{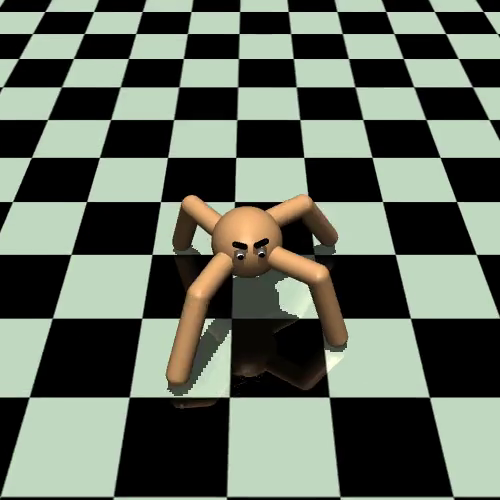}
  \vspace{0.2cm}
  \end{minipage}
  \begin{minipage}{0.19\linewidth}
  \includegraphics[width=0.95\linewidth]{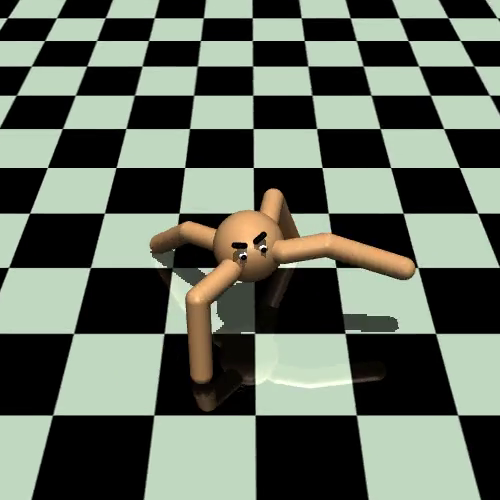}
  \vspace{0.2cm}
  \end{minipage}
  \begin{minipage}{0.19\linewidth}
  \includegraphics[width=0.95\linewidth]{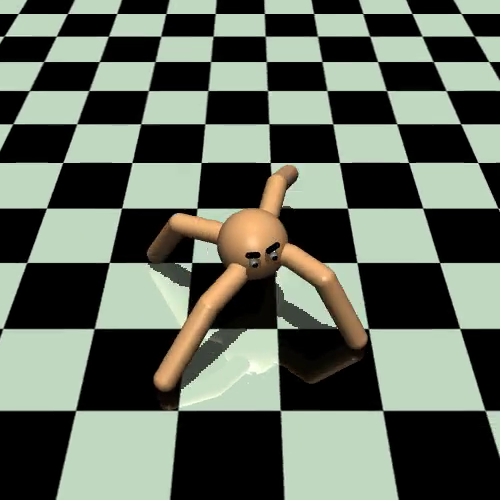}
  \vspace{0.2cm}
  \end{minipage}
  \begin{minipage}{0.19\linewidth}
  \includegraphics[width=0.95\linewidth]{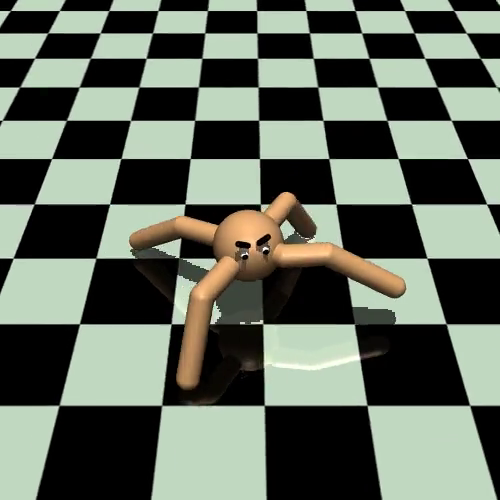}
  \vspace{0.2cm}
  \end{minipage}
  \begin{minipage}{0.19\linewidth}
  \includegraphics[width=0.95\linewidth]{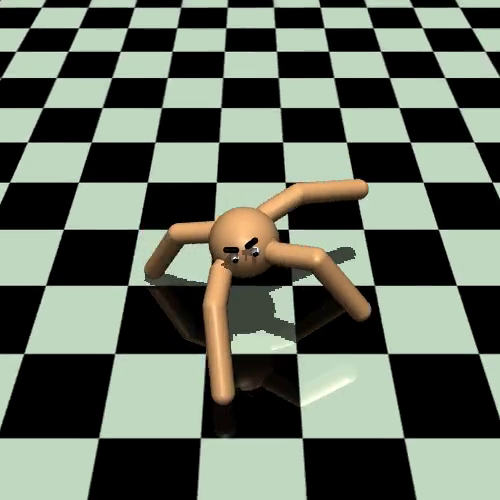}
  \vspace{0.2cm}
  \end{minipage} 
  \\
  \begin{minipage}{0.19\linewidth}
  \includegraphics[width=0.95\linewidth]{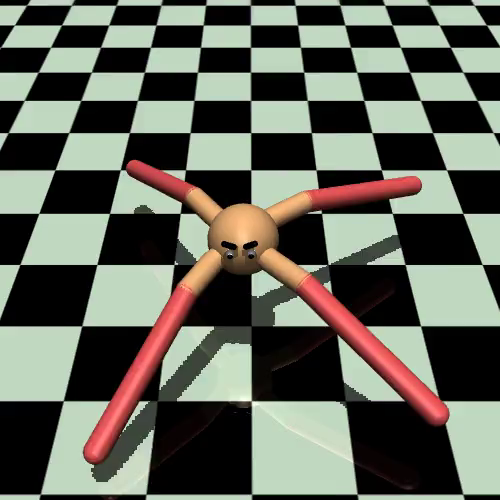}
  \vspace{0.2cm}
  \end{minipage}
  \begin{minipage}{0.19\linewidth}
  \includegraphics[width=0.95\linewidth]{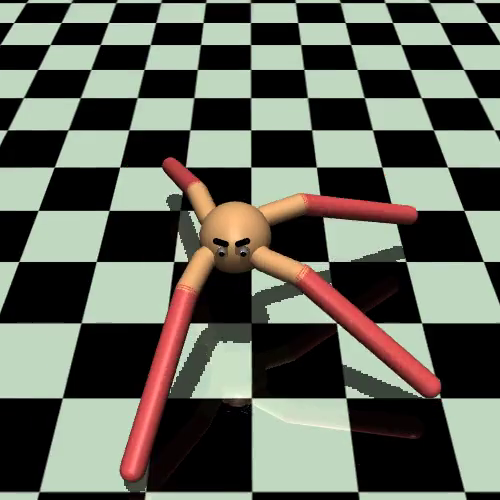}
  \vspace{0.2cm}
  \end{minipage}
  \begin{minipage}{0.19\linewidth}
  \includegraphics[width=0.95\linewidth]{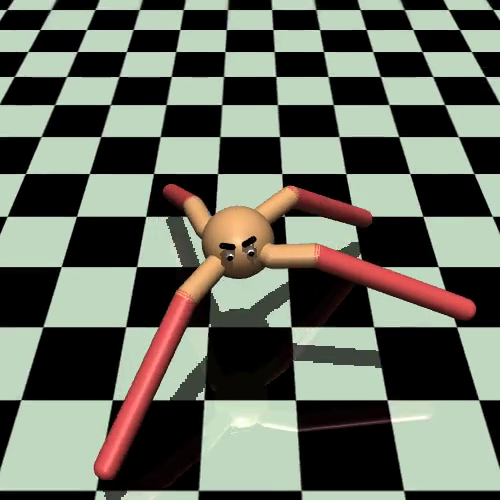}
  \vspace{0.2cm}
  \end{minipage}
  \begin{minipage}{0.19\linewidth}
  \includegraphics[width=0.95\linewidth]{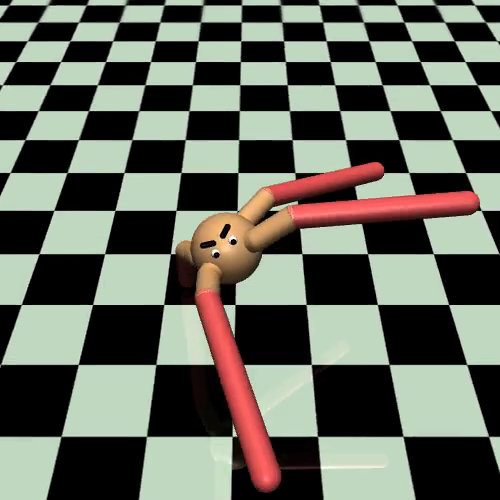}
  \vspace{0.2cm}
  \end{minipage}
  \begin{minipage}{0.19\linewidth}
  \includegraphics[width=0.95\linewidth]{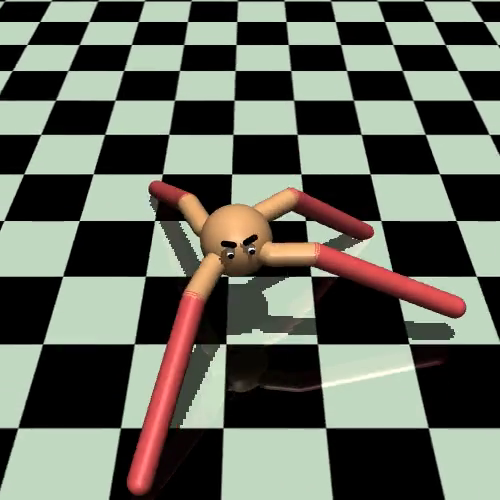}
  \vspace{0.2cm}
  \end{minipage}  
  \\
  \begin{minipage}{0.19\linewidth}
  \includegraphics[width=0.95\linewidth]{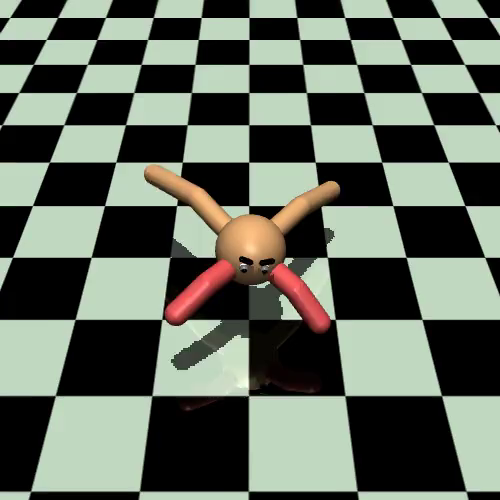}
  \vspace{0.2cm}
  \end{minipage}
  \begin{minipage}{0.19\linewidth}
  \includegraphics[width=0.95\linewidth]{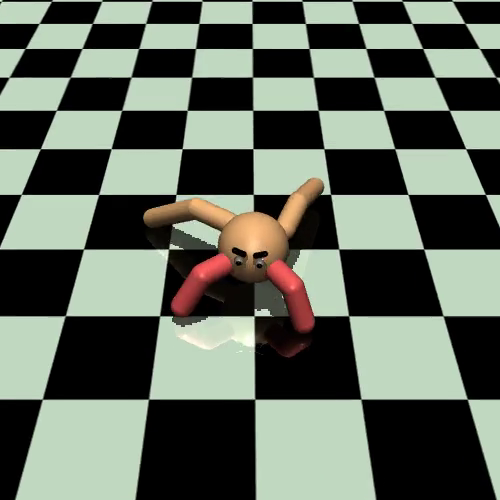}
  \vspace{0.2cm}
  \end{minipage}
  \begin{minipage}{0.19\linewidth}
  \includegraphics[width=0.95\linewidth]{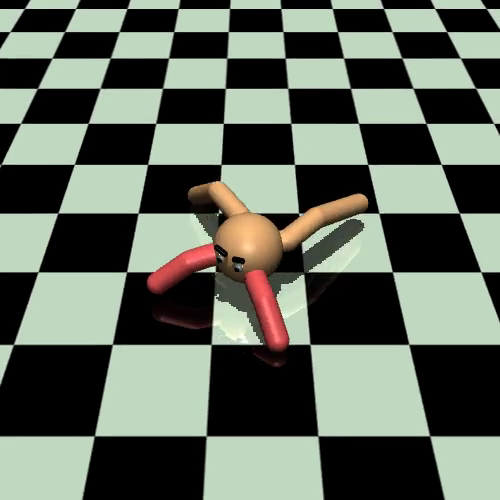}
  \vspace{0.2cm}
  \end{minipage}
  \begin{minipage}{0.19\linewidth}
  \includegraphics[width=0.95\linewidth]{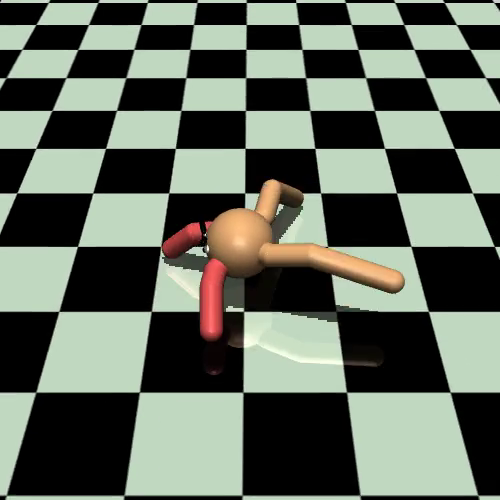}
  \vspace{0.2cm}
  \end{minipage}
  \begin{minipage}{0.19\linewidth}
  \includegraphics[width=0.95\linewidth]{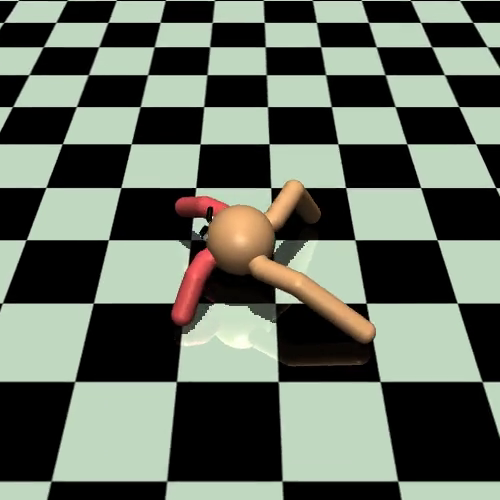}
  \vspace{0.2cm}
  \end{minipage}  
\end{tabular}
\caption{\textbf{Top row:} The original quadrupedal ant moving from left to right. \textbf{Middle row:} BigAnt with successfully learning the policy with transferred reward function from the original quadrupedal ant. \textbf{Bottom row:} AmputatedAnt moving from left to right. Due to the amputated legs, it cannot move like the other environments, instead it turns backward to move in the right direction.}
\label{fig:transfer_ant}
\end{figure*}

We visualize the movement of the agents on both source and taget environment in Fig.~\ref{fig:transfer_point_maze} and Fig.~\ref{fig:transfer_ant}.
In PointMaze environments, we can observe that both agents successfully learn to reach the goal in spite of differences in position of the border.
We further visualize the learned reward of OPIRL in Fig. \ref{fig:pointmaze_reward}.
We can see that the learned reward function generalizes well to the unseen environment (see Fig.~\ref{fig:transfer_laerning_environments}) and different goal positions.

Similarly, in quadrupedal Ant environment, BigAnt successfully learns to move from left to right in the same manner as the source environment.
However, due to two front legs being shortened, it is difficult for AmputatedAnt to move like the other models. 
Thus, it has to rotate to achieve smooth sideways movement.

\subsection{Additional Experiment Results}
We show additional information on sample efficiency comparison between IRL methods: OPIRL vs {\it f}-IRL.
We extract the learning curves of the two methods from Fig.~2 and plot the lines until it reaches the expert level for a couple of steps.
As shown in Fig.~\ref{fig:diff_firl} we can observe that our method significantly improves the sample efficiency for all tasks.
This is especially prominent on more complicated tasks such as Humanoid.

\begin{figure*}[tb]
    \centering
    \includegraphics[width=\linewidth]{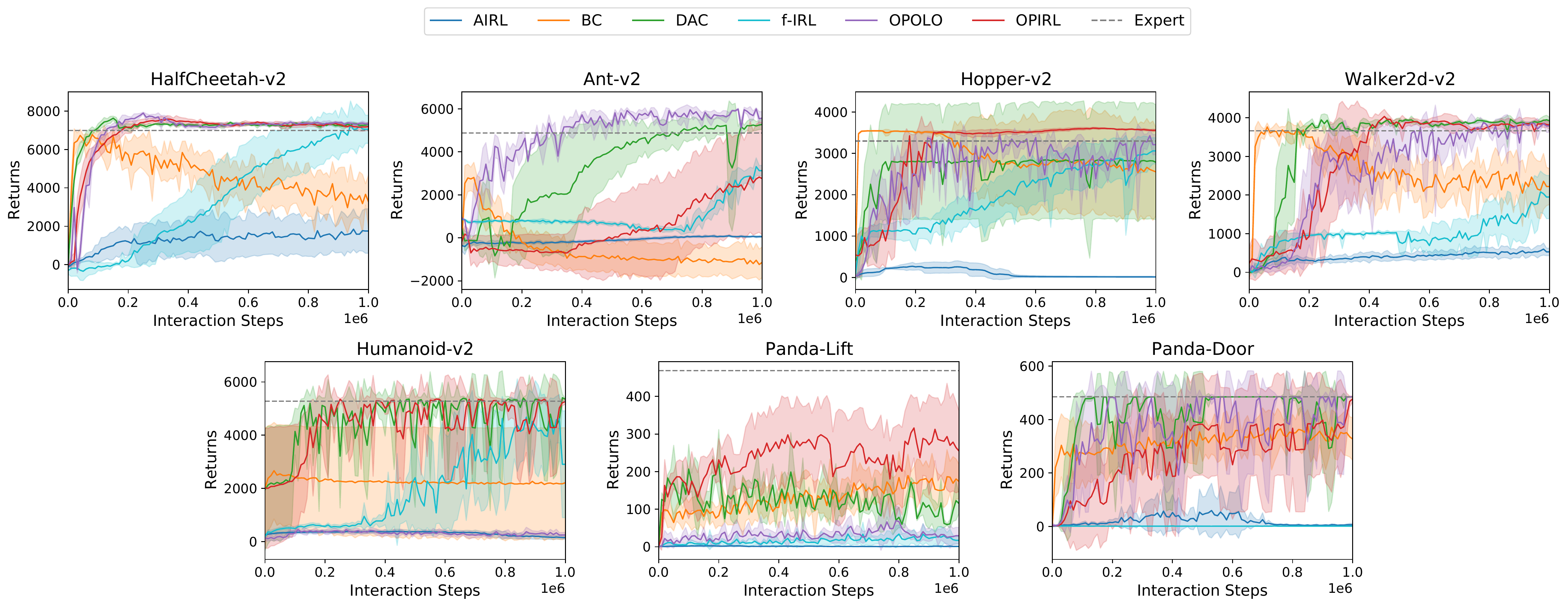}
    \caption{Comparison  between  the  average  return  of  the  trained  policy  vs  that  of  the  expert  policy  using 4 trajectory.  The  expert  policy performance is shown in a gray horizontal line, and we run each agent on 3 seeds and plot the mean and standard deviation.}
    \label{fig:results_4}
\end{figure*}

\begin{figure*}[tb]
    \centering
    \includegraphics[width=\linewidth]{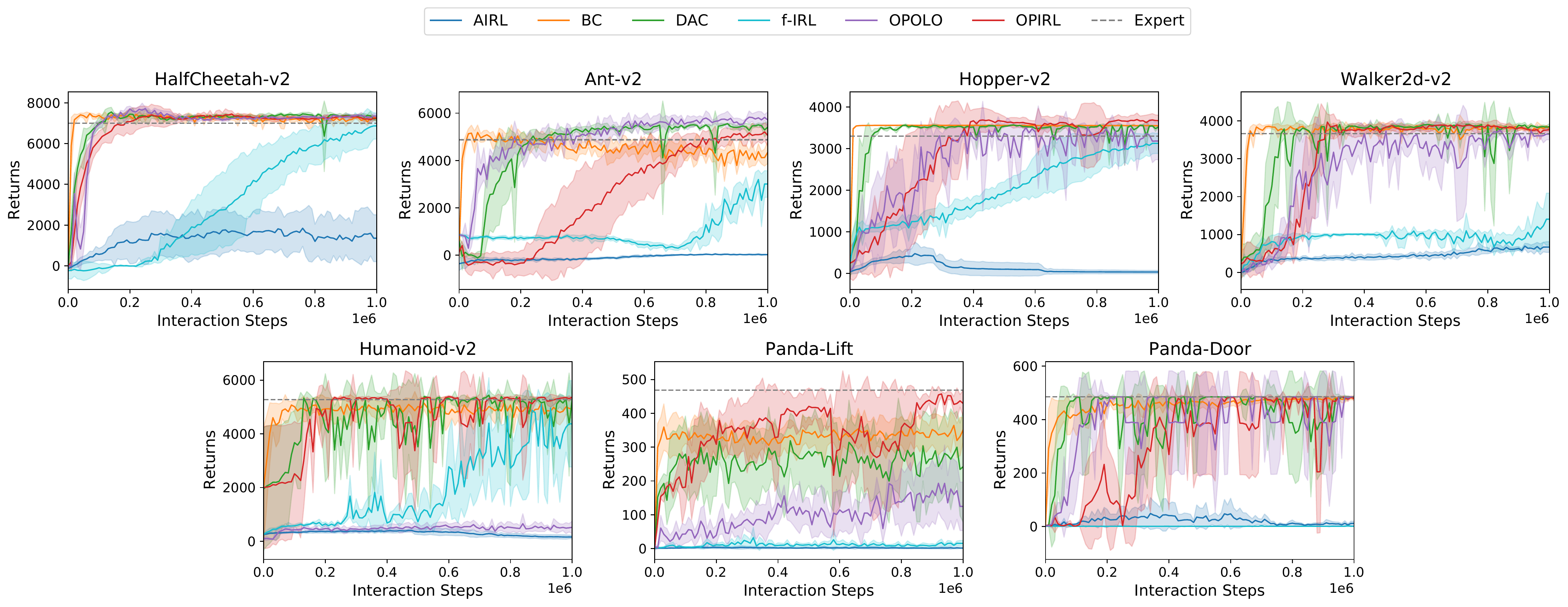}
    \caption{Comparison  between  the  average  return  of  the  trained  policy  vs  that  of  the  expert  policy  using 16 trajectory.  The  expert  policy performance is shown in a gray horizontal line, and we run each agent on 3 seeds and plot the mean and standard deviation.}
    \label{fig:results_16}
\end{figure*}

Furthermore, We show the learning curves for multiple trajectory results in Fig.~\ref{fig:results_4} and Fig.~\ref{fig:results_16}.
In all experiments, OPIRL DAC, and OPOLO show comparable results, except for OPIRL on Ant environment.
{\it f-IRL} shows decent performance on all tasks, however it cannot reach optimal level within 1M steps threshold.
BC performance increases as the number of trajectory grows.
This result is expected, and also seen in prior work \cite{ni_f-irl_2020}.
For AIRL, although tuning the learning rates, it failed on almost all tasks, except for HalfCheetah and Walker2D.

% %%%%%%%%%%%%%%%%%%%%%%%%%%%%%%%%%%%%%%%%%%%%%%%%%%%%%%%%%%%%%%%%%%%%%%%%%%%%%%%%
% \subfile{sections/appendix}
% %%%%%%%%%%%%%%%%%%%%%%%%%%%%%%%%%%%%%%%%%%%%%%%%%%%%%%%%%%%%%%%%%%%%%%%%%%%%%%%%
\end{document}